\documentclass[runningheads]{llncs}

\usepackage{eccv}
\usepackage{eccvabbrv}
\usepackage{graphicx}
\usepackage{booktabs}

\usepackage[accsupp]{axessibility} 

\usepackage{hyperref}
\usepackage{algorithm}
\usepackage{algpseudocode}
\usepackage{algorithmicx}
\usepackage{makecell}

\usepackage{amssymb}
\usepackage{gensymb}
\usepackage[utf8]{inputenc}
\usepackage[T1]{fontenc} 
\usepackage{url}
\usepackage{amsfonts} 
\usepackage{nicefrac} 
\usepackage{microtype} 
\usepackage{xcolor}

\usepackage{color}
\usepackage{amsmath}
\usepackage{mathtools}
\usepackage{arydshln}
\usepackage{bm}
\usepackage{enumitem}
\usepackage{multirow}
\usepackage{cuted}
\usepackage{tcolorbox}

\usepackage{orcidlink}

\newcommand{\bx}{\mathbf{X}}
\newcommand{\by}{\mathbf{Y}}
\newcommand{\dby}{\Delta \mathbf{Y}}
\newcommand{\bmask}{\mathbf{M}}
\newcommand{\be}{\mathbf{E}}

\begin{document}

\title{Accelerated Likelihood Maximization for Diffusion-based Versatile Content Generation} 

\titlerunning{Accelerated Likelihood Maximization for Versatile Content Generation}

\author{Hyunsoo Lee\inst{1}\orcidlink{0009-0000-3514-1602} \and
Inwoo Hwang\inst{1}\orcidlink{0009-0005-9819-1873} \and
Young Min Kim\inst{1,2\dagger}\orcidlink{0000-0002-6735-8539}}

\authorrunning{H.~Lee et al.}

\institute{ECE, Seoul National University \and
INMC \& IPAI, Seoul National University \\
\email{\{philip21,inusu0818,youngmin.kim\}@snu.ac.kr}}

\maketitle

\def\thefootnote{$\dagger$}\footnotetext{Corresponding author}


\begin{abstract}

Generating diverse, coherent, and plausible content from partially given inputs remains a fundamental challenge for diffusion models. 
Existing approaches face clear limitations: training-based approaches offer strong task-specific results but require costly computation, and they generalize poorly across tasks. 
Training-free approaches offer better efficiency, but they do not explicitly optimize over unobserved variables, leading to globally inconsistent results.
To address these limitations, we introduce Accelerated Likelihood Maximization (ALM), a novel training-free sampling strategy integrated into the reverse diffusion process that significantly extends the applicability of diffusion models beyond simple generation tasks. 
Unlike previous methods that implicitly influence missing regions through pre-generated region constraints, we directly optimize the unobserved region during the sampling process, enabling globally coherent and plausible generation.
Furthermore, we incorporate an acceleration strategy that significantly improves computational efficiency without sacrificing performance. 
Experimental results demonstrate that ALM consistently outperforms state-of-the-art methods in various data domains and tasks, establishing a powerful paradigm for versatile content generation.
Project website: \url{http://hleephilip.github.io/ALM}
\keywords{Diffusion Models \and Versatile Generation \and Synchronization}

\end{abstract}


\section{Introduction}
\label{sec:intro}

Diffusion models~\cite{ho2020denoising, song2020denoising, song2020score} have demonstrated remarkable performance in visual synthesis, including images~\cite{rombach2022high, saharia2022photorealistic, podell2023sdxl, peebles2023scalable, esser2024scaling, flux2024, wu2025qwenimagetechnicalreport}, videos~\cite{wang2025lavie, HaCohen2024LTXVideo, hong2022cogvideo, yang2024cogvideox}, and human motions~\cite{tevet2022human, karunratanakul2023guided} learned from massive datasets~\cite{schuhmann2022laion, kakaobrain2022coyo, bain2021frozen, guo2022generating}.
Within their training domains and task formulations, these models can produce highly realistic and semantically rich outputs. 
However, the generative power of the models is not directly applicable in practical scenarios that require generating content conditioned on partially observed or pre-generated inputs.
Exemplar scenarios include: filling in missing regions~\cite{lugmayr2022repaint, corneanu2024latentpaint, avrahami2023blended, ju2024brushnet, zhuang2024task, manukyan2023hd}, extrapolating beyond pre-generated boundaries~\cite{lee2023syncdiffusion, kim2024synctweedies, lee2025syncsde, yeo2025stochsync}, or lifting 2D image generation to view-consistent 3D texture generation~\cite{liu2024text, youwang2024paint, zeng2024paint3d, richardson2023texture, zhang2024texpainter, yan2025flexpainter}.
These real-world scenarios involve completion and extension tasks that go far beyond the standard generate-from-scratch setting.
We refer to this broader problem formulation as \textit{versatile content generation}.

A na\"ive approach is retraining or fine-tuning pre-trained models for each specific task, which requires substantial computational resources and large-scale dedicated datasets.
More importantly, these models are fundamentally limited in generalizability; 
they are trained for one specific task or modality and rarely transfer to others, even within the same modality.
However, although task-specific tuning is inefficient, the pretrained generative models themselves remain extremely powerful. 
This observation motivates our goal: 
we aim to transform the pretrained generative models from simple synthesis frameworks to versatile content generators, without any additional training.

With a shared motivation, diffusion synchronization~\cite{lee2023syncdiffusion, kim2024synctweedies, yeo2025stochsync, lee2025syncsde} attempts to address this problem. 
These methods are designed to achieve two goals: (a) fill the unobserved region while maintaining consistency with the pre-generated content, and (b) generate high-quality unobserved regions without sacrificing the original performance of the model.
However, existing methods are mostly heuristic or indirect: SyncTweedies~\cite{kim2024synctweedies} relies on extensive empirical search, while SyncSDE~\cite{lee2025syncsde} provides a mathematical justification but adopts a strong Gaussian assumption.
More importantly, SyncSDE's guidance is restricted to the pre-generated region and does not control the generation of unobserved regions.
The unobserved region remains unconstrained, under the assumption that completions will naturally emerge during the diffusion process. 
However, this assumption does not hold in practice, especially when the unobserved region becomes large (\textit{e.g.},  outpainting), thereby producing implausible outcomes. 
In essence, SyncSDE lacks a mechanism that enforces high-quality generation in the missing region, still failing to achieve the goal of synchronization.

To address these limitations, we propose \textit{Accelerated Likelihood Maximization (ALM)}, a fully \textbf{training-free} and \textbf{widely applicable} sampling mechanism by modifying the reverse diffusion sampling process.
ALM introduces a fundamentally different optimization principle from prior synchronization and posterior-guidance approaches~\cite{chung2022diffusion, song2023pseudoinverse, pandey2025variational, geyfman2026calibrated}. 
Instead of enforcing consistency only in pre-generated regions or sampling a single trajectory under posterior constraints, ALM is, to our knowledge, the first work to formulate synchronization as explicit likelihood maximization over the unobserved region to correlate multiple diffusion trajectories.
By directly performing likelihood maximization over the unobserved variable with respect to the pre-generated context and the diffusion prior, ALM improves the plausibility of missing regions while preserving global coherence.
This leads to a novel inference mechanism -- adaptive, region-aware likelihood maximization that directly updates the diffusion trajectories.
We use the term likelihood maximization in a score-based inference sense: the update follows score estimates of a composite log-probability objective over the unobserved variable while keeping the pretrained generative model fixed.
Technically, we derive an acceleration strategy that handles iterative likelihood maximization in a single step, significantly improving inference efficiency without sacrificing quality.

We demonstrate that ALM is broadly applicable across various generative models~\cite{rombach2022high, karunratanakul2023guided, zhang2023adding, wang2025lavie}, ranging from large-scale diffusion models (SDXL~\cite{podell2023sdxl}) to recent flow-matching frameworks (FLUX~\cite{flux2024}), enabling them to handle versatile generation tasks.
Further, ALM extends the application scope of pretrained models to include challenging tasks such as image inpainting, wide image generation, human motion completion, and 3D mesh texturing.
As a result, ALM establishes a general paradigm for versatile content generation. 
Below are our contributions:
\begin{itemize}[label=$\bullet$]
    \item We propose ALM, a fully training-free and model-agnostic sampling scheme that can directly transform pretrained generative models into versatile content generators.
    \item We formulate diffusion synchronization as explicit score-based likelihood maximization over the unobserved region, providing a principled optimization objective beyond prior synchronization-based approaches.
    \item We derive an accelerated one-step inference strategy and demonstrate broad applicability across images, human motion, 3D meshes, and videos, achieving state-of-the-art performance even compared to training-based baselines.
\end{itemize}


\section{Related Works}
\label{sec:rel_works}

\subsubsection{Training-based Methods.}
Several methods require training to address specific tasks of versatile content generation~\cite{ju2024brushnet, zhuang2024task, Liu_2026_CVPR, cohan2024flexible, yan2025flexpainter, liang2025UnitTEX, Lai_2026_CVPR, Wu_2026_CVPR}.
For image inpainting, BrushNet~\cite{ju2024brushnet} presents a plug-and-play dual-branch  architecture that separately processes masked image features from diffusion latents. 
Similarly, PowerPaint~\cite{zhuang2024task} introduces a framework with learnable task prompts, allowing a model to handle diverse inpainting challenges within the image domain.
There also exist variants of Stable Diffusion~\cite{rombach2022high} and SDXL~\cite{podell2023sdxl} that are specifically fine-tuned for image inpainting.
Beyond images, CondMDI~\cite{cohan2024flexible} extends diffusion models to human motion~\cite{tevet2022human} to perform human motion completion from partial keyframes, generating coherent and diverse motion sequences.
While these methods achieve strong performance on specific tasks, their reliance on extensive task-specific training limits their generalization across diverse domains, making them unsuitable for versatile content generation.
In contrast, ALM is fully training-free, shows broad generalizability across diverse tasks, and even outperforms training-based approaches.

\subsubsection{Training-free Methods.}
To overcome the high computational cost required for training-based methods, several task-specific training-free approaches~\cite{lugmayr2022repaint, manukyan2023hd, avrahami2023blended, ho2022video, liu2024text, richardson2023texture, zhang2024texpainter} have been proposed.
HD-Painter~\cite{manukyan2023hd} introduces prompt-aware attention and reweighted attention score guidance to guide the reverse diffusion process of inpainting models, combined with a tailored super-resolution module and Poisson blending~\cite{perez2023poisson}.
Reconstruction guidance~\cite{ho2022video}, originally proposed for long video generation, enforces consistency with pre-generated frames during denoising of the unobserved region using L2 loss.
This strategy can be extended to other modalities, such as human motion, as discussed in~\cite{cohan2024flexible}.
In the 3D domain, TexPainter~\cite{zhang2024texpainter} leverages the color blending scheme to ensure multi-view consistency during mesh texturing.
However, most of these techniques are tailored to specific tasks and therefore remain limited in applicability.
In contrast, we propose a unified framework that can be applied across modalities while achieving state-of-the-art performance.

\subsubsection{Diffusion Synchronization.}
Synchronization-based methods~\cite{bar2023multidiffusion, lee2023syncdiffusion, kim2024synctweedies, lee2025syncsde, yeo2025stochsync} propose tailored strategies to model the correlations between diffusion trajectories. 
For instance, SyncTweedies~\cite{kim2024synctweedies} evaluates 60 strategies and shows that the averaging variables obtained using Tweedie's formula yield the best results, though its effectiveness relies largely on heuristics without a clear mathematical explanation.
StochSync~\cite{yeo2025stochsync} relies on alternately sampled non-overlapping views, rather than explicitly modeling the relation between unobserved and pre-generated regions. 
Since it does not evaluate the interactions between trajectories within a single diffusion step, it inevitably results in global inconsistencies.
SyncSDE~\cite{lee2025syncsde} formulates the posterior distribution of the pre-generated content given the unobserved region, but still does not explicitly optimize the unobserved region and relies solely on guidance derived from the pre-generated content.
It implicitly assumes that plausible completions will emerge without directly enforcing them. 
Such an assumption lacks a solid foundation, and consequently it fails to generate high-quality content.
To overcome this limitation, we introduce a novel optimization objective that explicitly optimizes the unobserved region, thereby enhancing both consistency and overall fidelity.


\section{Proposed Method}
\label{sec:method}

\begin{figure}[t!]
	\centering
    \includegraphics[width=1.0\linewidth]{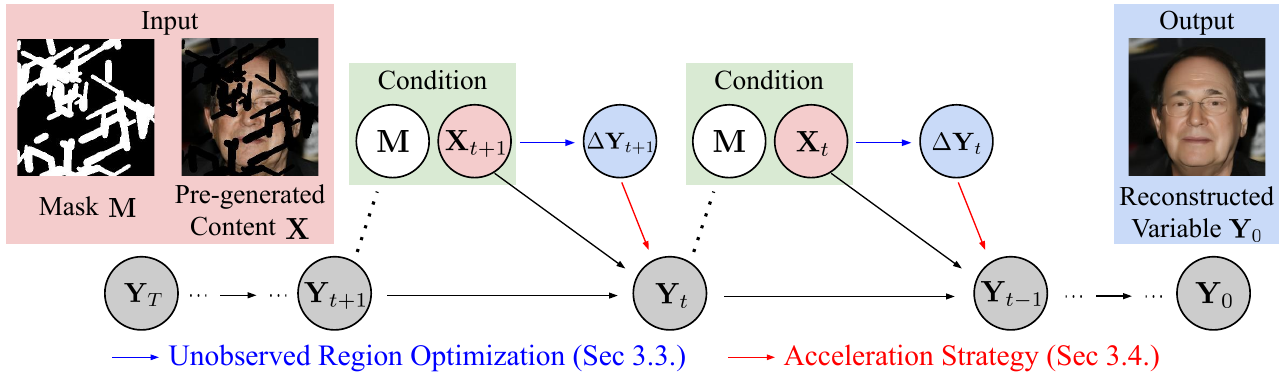}
	\vspace{-6mm}
	\caption{Overview of the proposed method. ALM aims to adapt diffusion models to reconstruct the unobserved variable while preserving pre-generated content. 
    }
	\vspace{-3mm}
\label{fig:method_overview}
\end{figure}

\subsection{Overview}
We aim to adapt diffusion-based generative models~\cite{sohl2015deep, ho2020denoising, song2020denoising, song2020score} to versatile inpainting and outpainting tasks in a training-free manner, where the unobserved variables are sampled while conditioning on the given pre-generated content.
We denote the pre-generated content as $\bx$ and the binary mask indicating the unobserved region as $\bmask$.
At diffusion timestep $t$, the noisy pre-generated content is represented as $\bx_t$, while the unobserved variable sampled by our method is written as $\by_t$.
We further define the blended variable $\be_t$ as $\be_t = \bx_t \odot (\mathbf{1}-\bmask) + \by_t \odot \bmask$.
During the reverse diffusion process, we first update $\by_t$ as follows:
\begin{equation}
    \by_t \leftarrow \by_t + \bmask \odot (w_1 (\epsilon_{\theta}(\by_t, t, \mathbf{c}) - \epsilon_{\theta}(\be_t, t, \mathbf{c})) - w_2 \epsilon_{\theta}(\be_t, t, \mathbf{c}) ),
    \label{eq:alm_main}
\end{equation}
where $\epsilon_{\theta}(\cdot, \cdot, \cdot)$ denotes the pretrained noise prediction network of the diffusion model, $\mathbf{c}$ is the conditioning variable (\textit{e.g.} text embedding), and $w_1, w_2$ are tunable hyperparameters.
Derivation of Eq.~\eqref{eq:alm_main} is the core contribution of our work.
We then sample $\by_{t-1}$ by modified DDIM~\cite{song2020denoising} reverse process as follows:
\begin{align}
    \by_{t-1} \leftarrow & \sqrt{\alpha_{t-1}} \left( \frac{\by_t - \sqrt{1-\alpha_t} \epsilon_{\theta}(\by_t, t, \mathbf{c})}{\sqrt{\alpha_t}}\right) + \sqrt{1 - \alpha_{t-1}} \epsilon_{\theta}(\by_t, t, \mathbf{c})  \nonumber   \\
    & +  w_1 (\mathbf{1}-\bmask)\odot (\bx_t - \by_t),
    \label{eq:syncsde_update_eq}
\end{align}
Figure~\ref{fig:method_overview} shows the graphical diagram of the proposed method.

\vspace{-2mm}

\subsection{Preliminary: SyncSDE}
\label{sec:method_prelimiary}
A representative approach tackling training-free versatile content generation is diffusion synchronization~\cite{kim2024synctweedies, yeo2025stochsync, lee2025syncsde}.
SyncSDE, which provides a probabilistic explanation of  why diffusion synchronization works, generates content by introducing a conditional probability term that couples different diffusion trajectories.
Specifically, it factorizes the conditional score function of the diffusion model used during the sampling of $\by_t$ as:
\begin{equation}
    \nabla_{\by_t} \log p(\by_t \mid \bx_t, \mathbf{c}) = \nabla_{\by_t} \log p(\by_t \mid \mathbf{c}) + \nabla_{\by_t} \log p(\bx_t \mid \by_t, \mathbf{c}),
\end{equation}
where the conditional probability of $\bx_t$ given $\by_t$ and $\mathbf{c}$ is modeled as:
\begin{equation}
    p(\bx_t \mid \by_t, \mathbf{c}) :=p(\bx_t \mid \by_t) \sim \mathcal{N}(\by_t, \frac{\gamma_t}{w_1} (1-\alpha_t) (\mathbf{1}-\bar{\bmask})^{-1}),
\end{equation}
with a diagonal precision matrix $\bar{\bmask}$, where pre-generated and unobserved entries are set to 0 and 1, respectively.
The conditional score is then substituted into the DDIM reverse process, yielding the update rule derived in  Eq.~\eqref{eq:syncsde_update_eq}.

\vspace{-2mm}

\subsection{Unobserved Region Optimization}
\label{sec:method_alm}

\subsubsection{Analysis.} 
The synchronization strategy discussed in Sec.~\ref{sec:method_prelimiary} often yields suboptimal results, since the guidance mechanism of SyncSDE~\cite{lee2025syncsde} focuses solely on optimizing the pre-generated region, $(\mathbf{1}-\bmask)\odot \by_t$, without explicitly providing any information for the unobserved region, $\bmask \odot \by_t$.
In other words, it does not guarantee that the unobserved region will be harmonized with the pre-generated content; 
instead, it just assumes that diffusion will naturally produce a plausible outcome, which is typically insufficient and does not hold.
To validate this analysis, we conduct an experiment on image inpainting.
As shown in Figure~\ref{fig:inpainting_cmp} (``SyncSDE'' column) and Figure~\ref{fig:ablation_study_qual} (1st row, ``w/o ALM'' columns), it often fails to synthesize coherent and high-quality outputs, where the unobserved regions  contain inconsistent or arbitrarily generated content that does not harmonize with the pre-generated region, supporting our analysis.

\subsubsection{Derivation.} 
We aim to optimize the unobserved region of $\by_t$ by imposing a novel sampling strategy.
At each diffusion timestep $t$, we introduce an additional term $\dby_t$, which is added to $\by_t$ for direct optimization.
We design $\dby_t= \sum_{i=1}^N {\dby_t^i}$, where the sequence $\{ \dby_t^i \}_{i=1}^N$ is constructed to iteratively minimize the following terms:
\begin{equation}
     -\lambda_1  \log  p(\bx_t , \bmask \mid \by_t^i + \bmask \odot \dby_t^i, \mathbf{c}) - \lambda_2  \log  p(\bx_t , \bmask , \by_t^i + \bmask \odot \dby_t^i \mid \mathbf{c}),
    \label{eq:f_dby}
\end{equation}
with $\lambda_1$ and $\lambda_2$ being scalar hyperparameters ($\lambda_1 > \lambda_2$).
Note that $\by_t^i = \by_t^{i-1} + \bmask \odot \dby_t^{i-1}$, and the initial values are set as $\by_t^1= \by_t$ and $\{\dby_t^i\}_{i=1}^N = \{  \mathbf{0}\}_{i=1}^N$.
Here, the conditional likelihood term encourages contextual consistency by aligning the unobserved region with the pre-generated content, whereas the joint log-density term encourages the blended content to lie within high-density regions of the full data distribution, thereby harmonizing both regions into a globally realistic sample.
We refer to the resulting procedure as likelihood maximization in a score-based sense, since the update follows score estimates of the corresponding composite log-probability objective over the unobserved region while keeping the pretrained model fixed.
This separation enables our method to simultaneously promote local consistency and global harmonization.
We verify that using both terms is essential for high-quality content generation in Sec.~\ref{sec:exp_img_inpainting}.
The coefficients $\lambda_1$ and $\lambda_2$ act as weights in a composite energy function~\cite{song2020score}, allowing adaptive balancing between two terms for better performance.

We define $f(\dby_t^i)$ as the objective defined in Eq.~\eqref{eq:f_dby}.
With the constraint of $\| \dby_t^i \|  \ll 1$, we apply a Taylor expansion around $\mathbf{0}$.
By taking a gradient descent on $\dby_t^i$ with step size of 1, we obtain:
\begin{equation}
    \dby_t^i =\bmask \odot (\lambda_1 \nabla_{\by_t^i} \log p(\bx_t, \bmask \mid \by_t^i, \mathbf{c}) + \lambda_2 \nabla_{\by_t^i} \log p(\bx_t, \bmask , \by_t^i \mid \mathbf{c})).
    \label{eq:optimization_objective}
\end{equation}
Note that the small-magnitude constraint can be satisfied by choosing sufficiently small values of $\lambda_1$ and $\lambda_2$, which we detail in Sec.~\ref{sec:method_acceleration}.
Using Bayes' rule, we factorize the conditional log-likelihood term into $p(\bx_t, \bmask , \by_t^i \mid \mathbf{c})$ and $p(\by_t^i \mid \mathbf{c})$. 
Following the score-based substitution technique~\cite{song2020score, lee2023conditional}, the second term is calculated using the pretrained diffusion model:
\begin{equation}
    \nabla_{\by_t^i} \log  p(\by_t^i \mid \mathbf{c}) \simeq - \frac{1}{\sqrt{1-\alpha_t}} \epsilon_{\theta}(\by_t^i, t, \mathbf{c})
    \label{eq:grad_log_y}
\end{equation}
For the first term, we define $\nabla_{\by_t^i} \log p(\bx_t, \bmask, \by_t^i \mid \mathbf{c}) \simeq \nabla_{\by_t^i} \log p(\be_t^i \mid \mathbf{c})$.
We justify that this score estimation works well in Appendix C, where it is interpreted as a surrogate for the joint score.
Then we get
\begin{equation}
    \nabla_{\by_t^i} \log p(\be_t^i \mid \mathbf{c}) = \nabla_{\be_t^i} \log p(\be_t^i \mid \mathbf{c}) \odot \bmask \simeq - \frac{1}{\sqrt{1-\alpha_t}}\epsilon_{\theta} (\be_t^i, t, \mathbf{c}) \odot \bmask.
    \label{eq:grad_log_e}
\end{equation}
Putting these together, the closed form formula for $\dby_t^i$ becomes
\begin{equation}
    \dby_t^i = \bmask \odot (\lambda_1 (\epsilon_{\theta}({\by}_t^i, t, \mathbf{c}) - \epsilon_{\theta}(\be_t^i ,t, \mathbf{c})) - \lambda_2 \epsilon_{\theta}(\be_t^i, t, \mathbf{c})),
    \label{eq:dby_closed_form}
\end{equation}
up to a scaling factor of $1/\sqrt{1-\alpha_t}$.

\vspace{-2mm}

\subsection{Acceleration Strategy}
\label{sec:method_acceleration}

From Eq.~\eqref{eq:dby_closed_form}, we can choose $\lambda_1$ and $\lambda_2$ such that each $\dby_t^i$ remains sufficiently small for accurate Taylor expansion, then get a sequence $\{ \dby_t^i \}_{i=1}^N$ with $N$ iterations.
However, this iterative process is computationally expensive, since its time complexity scales as $\mathcal{O}(N)$.
To address this, we propose a \textit{one-step approximation} strategy.
For the rest of the derivation, we denote $\by_t^i=\by_t^{i-1} + \dby_t^{i-1}$ from the definition of $\dby_t^{i-1}$.
Here, we present two claims for derivation:
\vspace{1mm}
\noindent \fbox{
\parbox{0.96\linewidth}{
    \textbf{Claim 1.} $\dby_t^i$ is small enough for all $1 \leq i \leq N$. 
    That is, $\lambda_1$ and $\lambda_2$ are chosen such that 
    $\| \dby_t^i \| \ll 1$. \\
    \textbf{Claim 2.} The noise prediction network $\epsilon_{\theta}(\cdot, \cdot, \cdot)$ of the pretrained diffusion model is $L$-Lipschitz~\cite{karras2022elucidating,kim2024fifo}.
    }
}
\newline
Using these claims, we analyze the difference between $\dby_t^i$ and $\dby_t^{i+1}$:
\begin{align}
    \|\dby^{i+1}_t  - \dby_t^{i} \| 
    \leq &  \; \lambda_1 \| \epsilon_{\theta}(\by_t^i + \dby_t^i , t, \mathbf{c}) - \epsilon_{\theta}(\by_t^i, t, \mathbf{c}) \|  \nonumber \\ 
    & + (\lambda_1 + \lambda_2 ) \| \epsilon_{\theta}(\be_t^i + \dby_t^i, t, \mathbf{c} ) - \epsilon_{\theta}(\be_t^i, t, \mathbf{c}) \| \nonumber \\
    \leq  & \; L(2 \lambda_1 + \lambda_2) \| \Delta \by_t^i \|
    =  \mathcal{O}(\| \Delta \by_t^i \|)
\end{align}
From Claim 1, it follows that $\dby_t^{i+1} \simeq \dby_t^i$ for all $i$.
Therefore, we approximate the iterative update with a one-step approximation as follows:
\begin{align}
    \dby_t \simeq N \dby^1_t 
    & = \bmask \odot (w_1' (\epsilon_{\theta}(\by_t, t, \mathbf{c}) - \epsilon_{\theta}(\be_t, t, \mathbf{c})) - w_2 \epsilon_{\theta}(\be_t, t, \mathbf{c}) ),
    \label{eq:acceleration_closed_form}
\end{align}
where we define $w_1' = N \lambda_1$ and $w_2 = N  \lambda_2$.
In practice, we set $w_1 = w_1'$, yielding only two hyperparameters.

We justify that when Claim 1 holds, $\| \dby_t^{i+1} - \dby_t^{i} \| \simeq 0$, making the one-step approximation valid in Appendix D.
Note that the use of $\mathcal{O}(\cdot)$ bounds for analyzing diffusion dynamics is not uncommon in the literature~\cite{kim2024fifo}, supporting the reasonableness of our derivation with strong empirical results.
Thanks to this approximation, instead of gradually refining the unobserved region through $N$ iterations, we directly compute the outcome of the full optimization in a single update.
This technique significantly reduces computation time \textit{without sacrificing the performance} as verified in Sec.~\ref{sec:exp_img_inpainting}.
In practice, we apply a decaying schedule to hyperparameters to better ensure the small-update assumption, defined as:
\begin{equation}
    w_i = \sigma_t w_i^{\mathrm{init}}, \quad \sigma_t = \sqrt{\frac{1-\alpha_{t-1}}{1-\alpha_t}} \sqrt{1 - \frac{\alpha_t}{\alpha_{t-1}}}
\end{equation}
where $\sigma_t$ follows the same definition as in DDPM~\cite{ho2020denoising}.

Additional methodological details are provided in the appendix.
Appendix A describes the full derivation of ALM, including the Taylor expansion and acceleration formulation.
Appendix C discusses the score estimation used to approximate the joint score, and Appendix D validates the one-step approximation.


\section{Experiments}
\label{sec:exp}

We comprehensively evaluate our approach on both inpainting and outpainting across diverse data modalities, highlighting its capability for versatile content generation. 
Specifically, we assess image inpainting in Sec.~\ref{sec:exp_img_inpainting} and wide image generation through outpainting in Sec.~\ref{sec:exp_wide_img_gen}.
Beyond the image domain, we extend our framework to human motion completion in Sec.~\ref{sec:exp_human_motion_completion}, and further explore its applicability to 3D mesh texturing in Sec.~\ref{sec:exp_mesh}.
For each table, we \textbf{bold} and \underline{underline} the best and second-best results, respectively.
Additional experimental material is provided in the appendix.
Appendix B provides task-specific details and additional results, including long video generation.
Appendix E analyzes computational cost, and Appendix F discusses hyperparameter sensitivity.

\subsection{Implementation Details}

We implement our method based on PyTorch~\cite{paszke2019pytorch}.
To ensure accurate score estimation in Eq.~\eqref{eq:grad_log_y} and Eq.~\eqref{eq:grad_log_e}, we do not employ classifier-free guidance~\cite{ho2022classifier} during the calculation of  Eq.~\eqref{eq:dby_closed_form}.
However, for fair comparison with baselines, we still apply classifier-free guidance in the reverse diffusion process of Eq.~\eqref{eq:syncsde_update_eq}.
For SyncSDE~\cite{lee2025syncsde}, since the official codebase does not support image inpainting scenarios, we reproduced it.
For all other baselines, we run the official codes of each algorithm for fair comparison.

\vspace{-2mm}

\subsection{Image Inpainting}

\label{sec:exp_img_inpainting}

\begin{figure}[b!]
	\centering
    \vspace{-6mm}
	\includegraphics[width=1.0\linewidth]{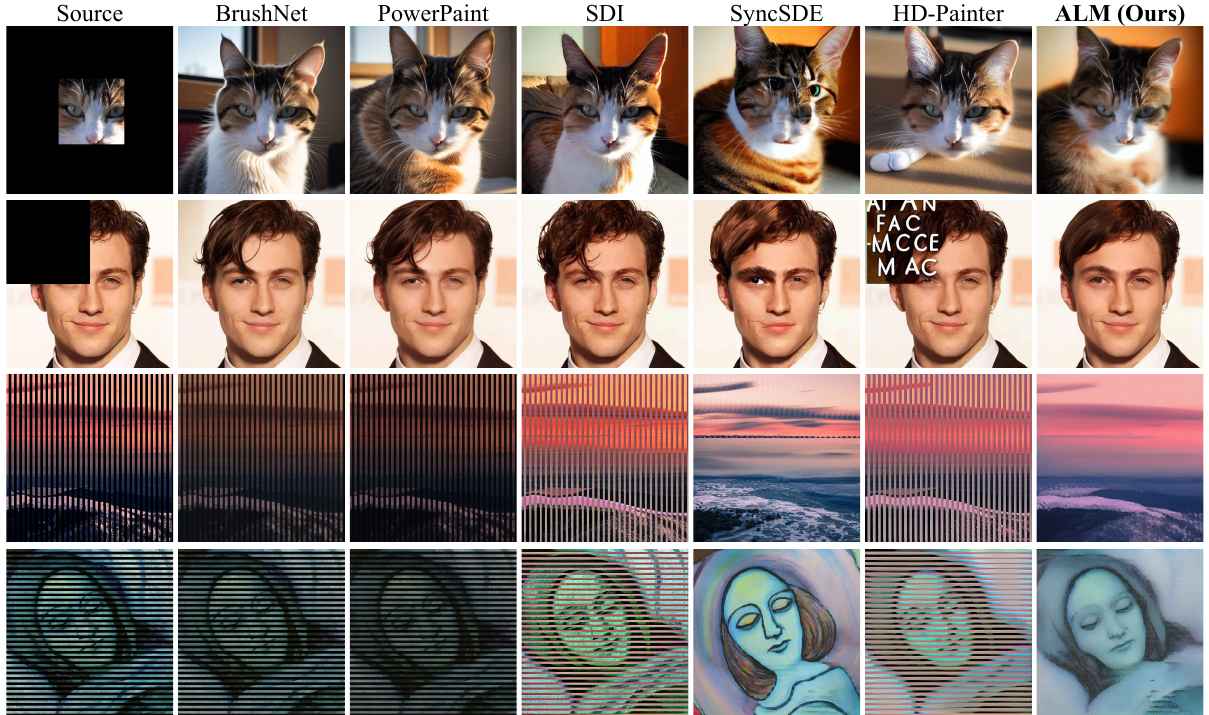}
	\vspace{-6mm}
	\caption{Qualitative comparison of our method against Stable Diffusion-based~\cite{rombach2022high} image inpainting methods~\cite{ju2024brushnet, zhuang2024task, rombach2022high, lee2025syncsde, manukyan2023hd}.
    ALM shows superior performance.}
\label{fig:inpainting_cmp}
\end{figure}

\subsubsection{Comparison with Prior Works.}

We first compare ALM against a wide range of representative inpainting methods using the pretrained Stable Diffusion~\cite{rombach2022high}.
The baselines include training-based methods built upon Stable Diffusion, such as BrushNet~\cite{ju2024brushnet}, PowerPaint~\cite{zhuang2024task}, and Stable Diffusion Inpainting (SDI)~\cite{rombach2022high}, as well as training-free methods such as SyncSDE~\cite{lee2025syncsde} and HD-Painter~\cite{manukyan2023hd}.
By comparing our method against SyncSDE~\cite{lee2025syncsde}, we provide strong supporting evidence for our analysis of SyncSDE's limitations presented in Sec.~\ref{sec:method_alm}.
We use two distinct datasets for evaluation: AFHQ~\cite{choi2020stargan} and CelebA-HQ~\cite{karras2017progressive}.
From each dataset, we sample 1,000 image--mask pairs to construct the test set.
We measure the performance using five commonly adopted metrics: LPIPS~\cite{zhang2018perceptual}, MSE, Masked SSIM (M-SSIM), Multi-Scale SSIM (MS-SSIM)~\cite{wang2003multiscale}, and FSIM~\cite{zhang2011fsim}.
Note that M-SSIM is computed over the unobserved region to better evaluate the quality of the generated region itself.
Since both ALM and SyncSDE require the sequence $\{ \bx_t \}_{t=0}^T$, we apply DDIM~\cite{song2020denoising} inversion with the masked source image.
For all algorithms, the masked source image is provided as input and the model generates the target image from pure noise, which follows the standard inpainting setup.

\begin{table}[t!]
\centering
\caption{Quantitative evaluation on image inpainting. Methods with * and $^\dagger$ denote results obtained with pixel-level blending and super-resolution, respectively.}
\vspace{-4mm}
\setlength{\tabcolsep}{3pt} 
\scalebox{0.63}{
\begin{tabular}{l c c c c c c c c c c c}
\toprule
\multirow{2}{*}{Method} & \multirow{2}{*}[-0.6ex]{\makecell[c]{Training-\\free}} & \multicolumn{5}{c}{AFHQ~\cite{choi2020stargan}} & \multicolumn{5}{c}{CelebA-HQ~\cite{karras2017progressive}} \\
\cmidrule(lr){3-7} \cmidrule(lr){8-12}
 & & LPIPS $\downarrow$ & MSE $\downarrow$ & M-SSIM $\uparrow$ & MS-SSIM $\uparrow$ & FSIM $\uparrow$ & LPIPS $\downarrow$ & MSE $\downarrow$ & M-SSIM $\uparrow$ & MS-SSIM $\uparrow$ & FSIM $\uparrow$   \\
\midrule
BrushNet~\cite{ju2024brushnet}     & N &  0.316 & 0.216 & 0.256 & 0.589 & 0.741 &   0.274 & 0.195 & 0.347 & 0.638 & 0.759 \\
PowerPaint~\cite{zhuang2024task}   & N &  0.310 & 0.217 & 0.272 & 0.600 & 0.748  &  0.272 & 0.203 & 0.366 & 0.650 & 0.764 \\
SDI~\cite{rombach2022high}          & N &  \underline{0.292} & \textbf{0.140} & 0.295 & 0.623 & 0.757  &  \underline{0.268} & \underline{0.130} & \underline{0.368} & 0.659 & 0.763 \\
SyncSDE~\cite{lee2025syncsde}      & Y &  0.304 & 0.172 & \underline{0.302} & \underline{0.641} & \underline{0.778}  &  0.292 & 0.159 & 0.341 & \underline{0.661} & \underline{0.781} \\
HD-Painter~\cite{manukyan2023hd}   & Y &  0.301 & 0.146 & 0.285 & 0.610 & 0.741  &   0.286  & 0.146 & 0.344 & 0.642 & 0.747 \\
ALM (Ours)   & Y &  \textbf{0.283} & \underline{0.143} & \textbf{0.351} & \textbf{0.689} & \textbf{0.796}   &  \textbf{0.251} & \textbf{0.126}  & \textbf{0.417} & \textbf{0.732} & \textbf{0.813} \\
\midrule
BrushNet*~\cite{ju2024brushnet}    & N &  0.286 & 0.201 & 0.270 & 0.622 & \underline{0.765}   &  \underline{0.250} & 0.183 & \underline{0.360} & 0.664 & \underline{0.779} \\
HD-Painter*$^\dagger$~\cite{manukyan2023hd} & Y &  \underline{0.285} & \underline{0.136} & \underline{0.300} & \underline{0.646} & \underline{0.765}   &  0.275 & \underline{0.140} & 0.347 & \underline{0.666} & 0.766   \\
ALM* (Ours)  & Y &  \textbf{0.259}  &  \textbf{0.125}  & \textbf{0.343} & \textbf{0.719} & \textbf{0.826}&   \textbf{0.240}  & \textbf{0.112} & \textbf{0.398} & \textbf{0.746} & \textbf{0.832}    \\
\bottomrule
\end{tabular}
}
\vspace{-4mm}
\label{tab:image_inpainting_cmp}
\end{table}

As summarized in Table~\ref{tab:image_inpainting_cmp}, our method demonstrates outstanding performance across all baselines. 
Notably, it consistently outperforms both training-free and training-based methods, regardless of the dataset.
We also emphasize that our method achieves better M-SSIM with a significant margin, showing the effects of explicit optimization of the unobserved region.
Figure~\ref{fig:inpainting_cmp} visualizes the qualitative comparisons, where our method consistently delivers superior visual quality. 
Our method also shows robust performance under diverse and complex mask geometries, demonstrating its generalizability.

\vspace{-2mm}

\subsubsection{Experiments across Diverse Backbones.}
To validate the robustness of ALM with respect to the underlying diffusion backbone, we conduct experiments with diverse models.
We adopt several additional backbones:
(a) an unconditional diffusion model trained on CelebA-HQ~\cite{karras2017progressive} from RePaint~\cite{lugmayr2022repaint},
(b) Stable Diffusion XL~\cite{podell2023sdxl}, and (c) FLUX~\cite{flux2024}.
FLUX uses flow matching~\cite{lipman2022flow, liu2022flow}, and we detail the adaptation of our method to the flow matching framework in Appendix B.1.
As shown in Figure~\ref{fig:inpainting_backbone}, our method delivers high-quality inpainting results on every backbone.
These findings highlight that ALM is model-agnostic, suggesting that our method transforms diverse pretrained models into versatile content generators without retraining, strengthening its importance.

\begin{figure}[h!]
	\centering
	\includegraphics[width=1.0\linewidth]{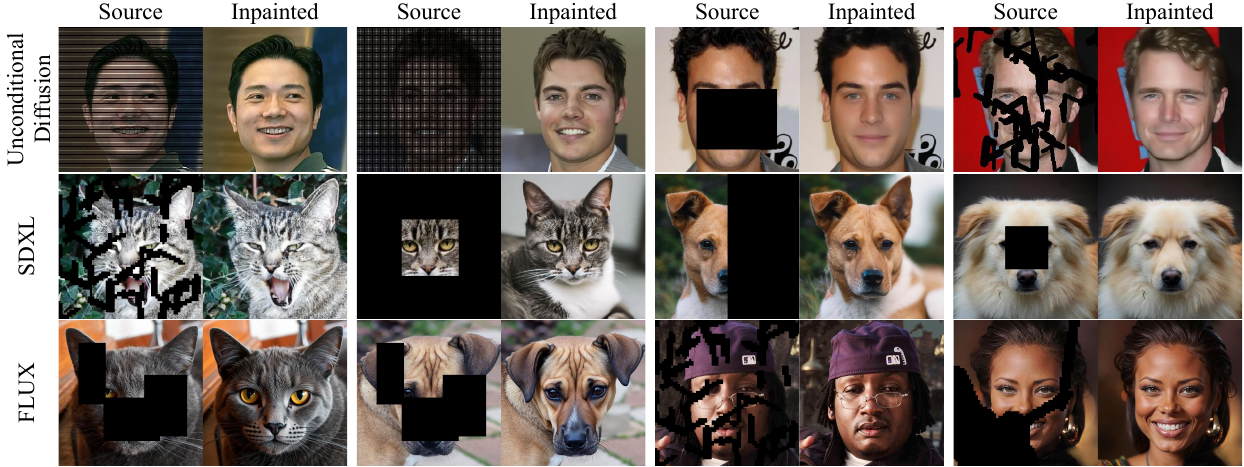}
	\vspace{-6mm}
	\caption{Qualitative results of image inpainting with various backbones~\cite{lugmayr2022repaint,podell2023sdxl, flux2024}.}
    \vspace{-4mm} 
\label{fig:inpainting_backbone}
\end{figure}

\subsubsection{Comparison with Large-Scale Models.}

To further validate the effectiveness of ALM against large-scale generative models, we compare it with SDXL-Inpainting~\cite{podell2023sdxl}, as well as FLUX~\cite{flux2024} and Qwen-Image~\cite{wu2025qwenimagetechnicalreport} adapted for inpainting via latent blending. 
Since these models are designed for high-resolution generation, we adopt SDXL as the backbone for ALM to ensure a fair comparison at 1K resolution.
As shown in Table~\ref{tab:image_inpainting_cmp_foundation}, ALM not only outperforms but also achieves this without additional training, underscoring the significance of our approach.

\vspace{-4mm}
\begin{table}[h!]
\centering
\caption{Quantitative comparison of ALM on image inpainting with large-scale generative models~\cite{podell2023sdxl, flux2024, wu2025qwenimagetechnicalreport}.}
\vspace{-3mm}
\setlength{\tabcolsep}{2.5pt} 
\scalebox{0.6}{
\begin{tabular}{l c c c c c c c c c c c}
\toprule
\multirow{2}{*}{Method}  & \multirow{2}{*}[-0.6ex]{\makecell[c]{Training-\\free}} & \multicolumn{5}{c}{AFHQ~\cite{choi2020stargan}} & \multicolumn{5}{c}{CelebA-HQ~\cite{karras2017progressive}} \\
\cmidrule(lr){3-7} \cmidrule(lr){8-12}
& & LPIPS $\downarrow$ & MSE $\downarrow$ & M-SSIM $\uparrow$ & MS-SSIM $\uparrow$ & FSIM $\uparrow$ & LPIPS $\downarrow$ & MSE $\downarrow$ & M-SSIM $\uparrow$ & MS-SSIM $\uparrow$ & FSIM $\uparrow$   \\
 \midrule
SDXL-Inpainting~\cite{podell2023sdxl}    & N & 0.303 & 0.191 & 0.309 & 0.656 & 0.780  &  0.313 &  0.232 &   0.332 &  0.657 &  0.768  \\
FLUX-Inpainting~\cite{flux2024}  & Y   &    0.272 & 0.155 & 0.290 & 0.662 & 0.788 & \textbf{0.212} & \underline{0.133} & \underline{0.364} & \underline{0.719} & \underline{0.809}
\\
Qwen-Image-Inpainting~\cite{wu2025qwenimagetechnicalreport}  & Y   &   \underline{0.257} & \underline{0.126} & \underline{0.333}  & \underline{0.689} & \underline{0.813} &  \underline{0.232} &  0.141 &  0.343 &  0.688 &  0.799  \\
ALM (Ours)  & Y  &  \textbf{0.254} & \textbf{0.112} &  \textbf{0.410} & \textbf{0.754} & \textbf{0.841} &  0.249 & \textbf{0.130} &  \textbf{0.442} & \textbf{0.772} & \textbf{0.843} \\
\bottomrule
\end{tabular}
}
\vspace{-10mm}
\label{tab:image_inpainting_cmp_foundation}
\end{table}

\subsubsection{Analyzing the Effect of Components.}

We analyze the effect of each objective term described in Eq.~\eqref{eq:f_dby}, as well as the overall impact of ALM and the acceleration strategy.
As shown in Table~\ref{tab:ablation}, each component plays an important role in generating high-quality results.
In particular, our one-step approximation does not degrade performance, while reducing the runtime by approximately 185$\times$.
This shows that the acceleration strategy greatly improves efficiency without sacrificing quality.
The exact runtime is reported in Appendix E.

\begin{figure}[t!]
	\centering
    \includegraphics[width=1.0\linewidth]{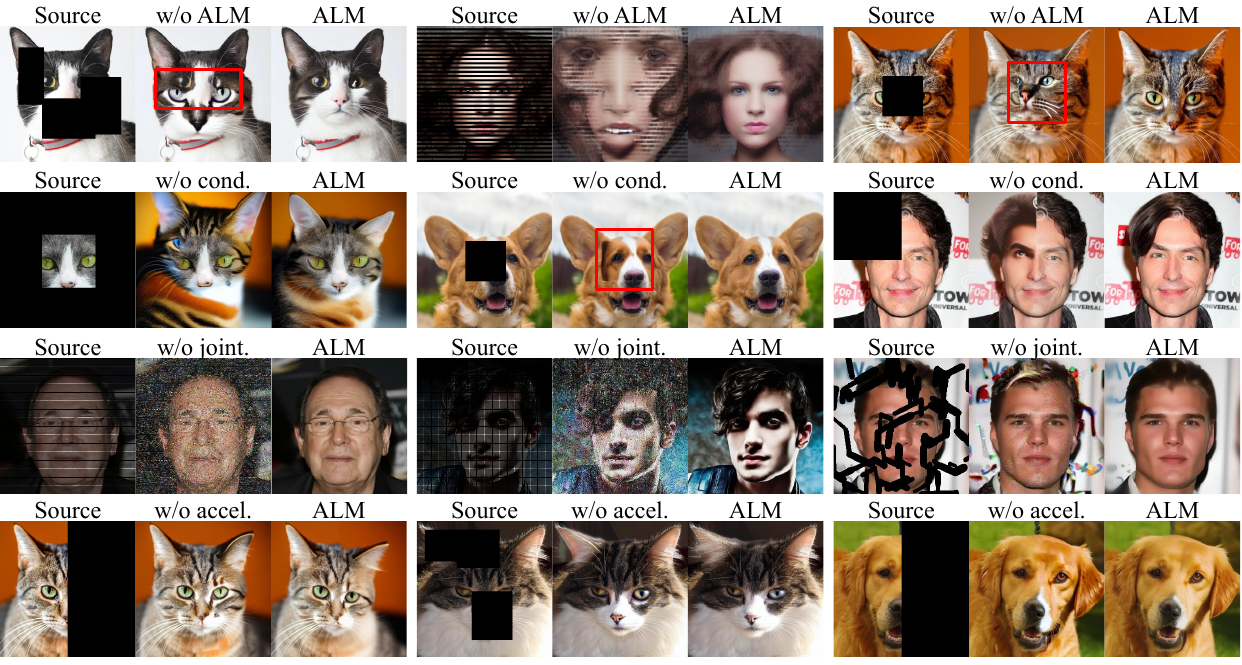}
	\vspace{-5mm}
	\caption{Qualitative ablation study results. Each row shows the effect of ALM (Eq.~\ref{eq:dby_closed_form}), the conditional likelihood term (Eq.~\ref{eq:f_dby}), the joint log-density term (Eq.~\ref{eq:f_dby}), and the acceleration strategy (Eq.~\ref{eq:acceleration_closed_form}), respectively.}
	\vspace{-5mm}
\label{fig:ablation_study_qual}
\end{figure}

We further present qualitative ablation results in Figure~\ref{fig:ablation_study_qual}.
The first row demonstrates that our method effectively mitigates a key limitation of the existing approach~\cite{lee2025syncsde}, which fails to harmonize the unobserved region with the pre-generated content.
In contrast, by explicitly performing likelihood maximization over the unobserved region, our approach produces globally coherent samples that align well with the given context.
The second and third rows visualize the effectiveness of the conditional likelihood and joint log-density terms, respectively.
Specifically, the conditional likelihood is more effective when applied to pretrained Stable Diffusion~\cite{rombach2022high}, whereas the joint log-density term has a significant impact on an unconditional diffusion model~\cite{lugmayr2022repaint}.
These results demonstrate that incorporating both terms enables ALM to generalize effectively across diverse diffusion backbones.
The last row shows that the visual quality remains consistent irrespective of the use of the acceleration strategy.

\begin{table}[h!]
\centering
\caption{Quantitative ablation study results on image inpainting.}
\vspace{-3mm}
\setlength{\tabcolsep}{3pt} 
\scalebox{0.67}{
\begin{tabular}{l c c c c c c c c c c}
\toprule
\multirow{2}{*}{Method}  & \multicolumn{5}{c}{AFHQ~\cite{choi2020stargan}} & \multicolumn{5}{c}{CelebA-HQ~\cite{karras2017progressive}} \\
\cmidrule(lr){2-6} \cmidrule(lr){7-11}
 & LPIPS $\downarrow$ & MSE $\downarrow$ & M-SSIM $\uparrow$ & MS-SSIM $\uparrow$ & FSIM $\uparrow$ & LPIPS $\downarrow$ & MSE $\downarrow$ & M-SSIM $\uparrow$ & MS-SSIM $\uparrow$ & FSIM $\uparrow$   \\
\midrule
w/o ALM     &   0.295 & 0.169 & 0.300 & 0.650 & 0.782 &  0.287 & 0.161 & 0.332 & 0.663 & 0.782 \\
w/o cond. term     &     0.295 & 0.162 &  0.323 & 0.661 & 0.787 &  0.291 & 0.156 & 0.357 &  0.673 & 0.786 
\\
w/o joint term     &   \underline{0.284} & \underline{0.149} & \underline{0.327} & \underline{0.679} & \underline{0.793} &  \underline{0.254} & \underline{0.132} & \underline{0.385} & \underline{0.720} & \underline{0.808} \\
w/o acceleration     &  0.298 & 0.170 & 0.298 & 0.649 & 0.781 &  0.277 & 0.156 & 0.341 & 0.675 & 0.787 \\
ALM (Ours)     &   \textbf{0.283} & \textbf{0.143} & \textbf{0.351} & \textbf{0.689} & \textbf{0.796} &  \textbf{0.251} & \textbf{0.126} & \textbf{0.417} & \textbf{0.732} & \textbf{0.813} \\
\bottomrule
\end{tabular}
}
\vspace{-2mm}
\label{tab:ablation}
\end{table}

\subsection{Wide Image Generation}
\label{sec:exp_wide_img_gen}
\vspace{-2mm}
Beyond image inpainting, our approach also naturally extends to the outpainting task, enabling the synthesis of wide, high-resolution images.
We employ an autoregressive image outpainting strategy to generate wide images. 
Starting from $512 \times 512$ patch generated with the pretrained Stable Diffusion~\cite{rombach2022high}, subsequent overlapping patches are iteratively synthesized via outpainting.
With a stride of 384 pixels, we generate five patches, resulting in a $2048 \times 512$ resolution image.
The patches are overlapped such that the $i$-th patch is placed on top of the $(i+1)$-th one and decoded with the pretrained VAE~\cite{kingma2013auto} decoder. 
We compare our method against state-of-the-art diffusion synchronization approaches using 400 images, including SyncTweedies~\cite{kim2024synctweedies},  SyncSDE~\cite{lee2025syncsde}, and StochSync~\cite{yeo2025stochsync}.
For evaluation, we randomly crop the generated wide image into $512 \times 512$ image.
We report FID~\cite{heusel2017gans} and KID~\cite{binkowski2018demystifying} to evaluate distribution alignment, with Aesthetic Score~\cite{schuhmann2022laion} and Q-Align~\cite{wu2023q} to assess the perceptual quality and fidelity of the generated images.

\begin{figure}[h!]
	\centering
    	\includegraphics[width=1.0\linewidth]{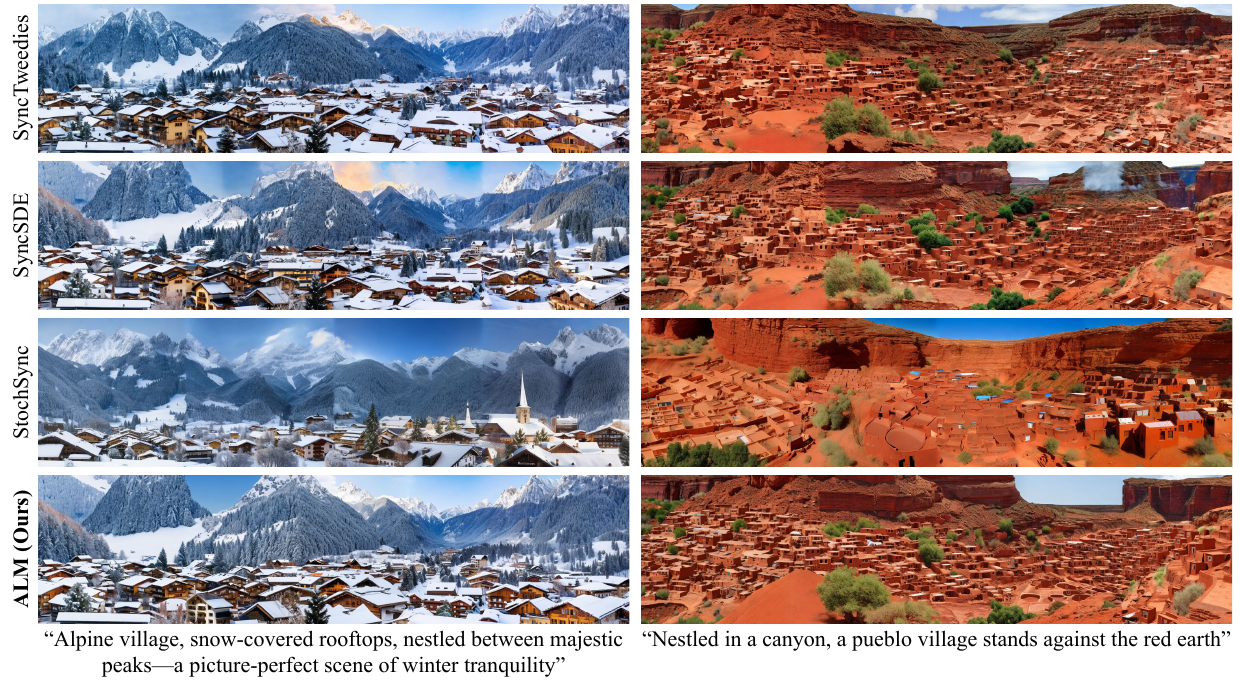}
	\vspace{-4mm}
	\caption{Qualitative comparison of our method against state-of-the-art methods~\cite{kim2024synctweedies, lee2025syncsde, yeo2025stochsync}.
    SyncSDE exhibits noticeable discontinuities across patches, while SyncTweedies produces blurry and inconsistent coloring. 
    StochSync likewise tends to generate blurry and discontinuous wide images. 
    In contrast, ALM successfully produces results without blurred or inconsistent regions.
    }
	\vspace{-10mm}
\label{fig:wide_image}
\end{figure}

As shown in Table~\ref{tab:wide_image_gen_cmp} and Figure~\ref{fig:wide_image}, our method achieves outstanding performance compared to the baselines.
In particular, SyncSDE (Row 2) exhibits clear limitations in the wide image generation setting. 
Since it does not explicitly optimize the unobserved region, it achieves inferior aesthetic quality in terms of aesthetic score and Q-Align.
Furthermore, since its performance degrades significantly when the masked area becomes large, it requires a smaller stride between overlapping patches. 
This leads to color inconsistencies and clear edge artifacts.

\vspace{-4mm}
\begin{table}[h!]
\caption{Quantitative evaluation on wide image generation.
KID~\cite{binkowski2018demystifying} is scaled by $10^3$.}
\centering
\setlength{\tabcolsep}{10pt} 
\vspace{-3mm}
\scalebox{0.9}{
    \begin{tabular}{l c c c c}
    \toprule
    Method & FID $\downarrow$ & KID  $\downarrow$  & Aesthetic Score $\uparrow$  &  Q-Align $\uparrow$   \\
    \midrule
    SyncTweedies~\cite{kim2024synctweedies} & 85.95 & 58.36 & 6.104   & \underline{4.550} \\
    SyncSDE~\cite{lee2025syncsde}      &  \underline{85.82} & \underline{51.84} & \underline{6.127} & 4.542  \\
    StochSync~\cite{yeo2025stochsync}      &  113.21 & 92.10 & 6.026  & 4.546 \\
    ALM (Ours)   &  \textbf{83.41} & \textbf{42.98} & \textbf{6.133} & \textbf{4.581} \\
    \bottomrule
    \end{tabular}
    }
    \vspace{-6mm}
    \label{tab:wide_image_gen_cmp}
\end{table}

\vspace{-6mm}

\subsection{Human Motion Completion}
\label{sec:exp_human_motion_completion}

We further demonstrate the versatility of our method by extending it from images to human motion data. 
Specifically, we tackle the human motion completion task, where the goal is to reconstruct missing parts of a motion sequence.
We evaluate across three distinct scenarios: ``first-half prediction,'' predicting the initial segment from the latter half, ``middle-half prediction,'' completing the central segment given the first and last quarters, and ``last-half prediction,'' the inverse of the first-half setting.
We use a U-Net-based~\cite{ronneberger2015u} pretrained diffusion model~\cite{karunratanakul2023guided} for text-to-motion synthesis.

\begin{figure}[h!]
    \vspace{-2mm}
	\centering
	\includegraphics[width=1.0\linewidth]{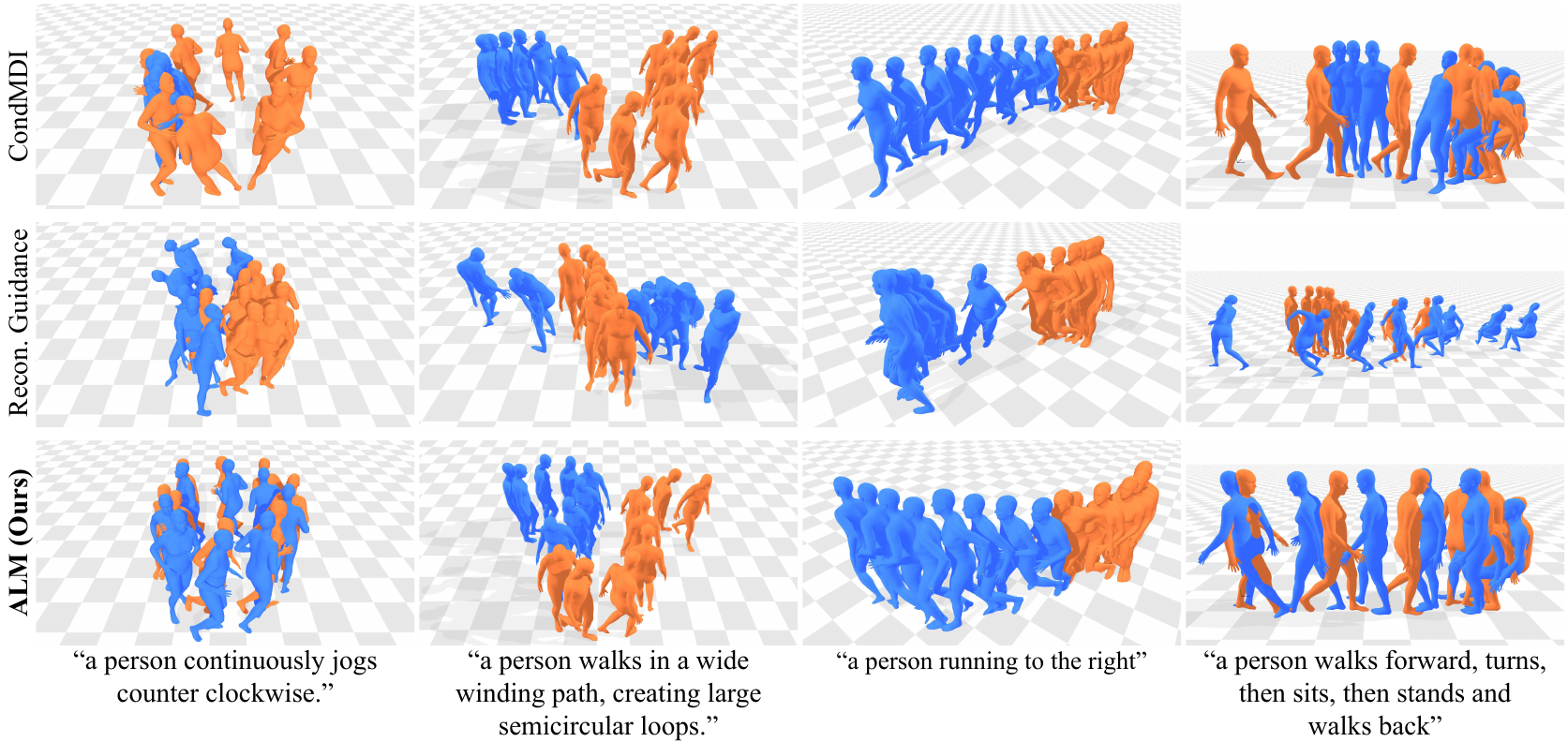}
	\vspace{-6mm}
	\caption{Qualitative comparison of our method with baselines~\cite{cohan2024flexible, ho2022video} on human motion completion. 
    While baselines show unrealistic or discontinuous motions, ALM generates plausible sequences that also align with the given text prompt.}
	\vspace{-10mm}
\label{fig:human_motion_completion}
\end{figure}

We compare our method against the training-based method CondMDI~\cite{cohan2024flexible} and training-free methods such as Reconstruction Guidance~\cite{ho2022video} and its imputation-based variant~\cite{tevet2022human}.
Basically, we follow the CondMDI setup that employs a DDPM~\cite{ho2020denoising} sampler with 1,000 timesteps.
For each completion scenario, we sample 1,000 motion sequences from the HumanML3D~\cite{guo2022generating} dataset and report the average performance over 10 replications.
We measure FID, Matching Score (Match.), Top-1 R-precision (R-prec.), and Diversity (Div.) metrics, which are widely adopted metrics in prior works~\cite{guo2022generating, cohan2024flexible}.
Table~\ref{tab:motion_completion_cmp} illustrates that the proposed method achieves superior performance with high versatility across various human motion completion scenarios. 
In Figure~\ref{fig:human_motion_completion}, the given frames are highlighted in \textcolor{orange}{orange}, while the filled frames generated by the model are shown in \textcolor{blue}{blue}.

\begin{table}[h!]
\vspace{-2mm}
\caption{Quantitative evaluation on human motion completion using the motion sequences sampled from HumanML3D~\cite{guo2022generating} dataset.
Methods marked with * use imputation~\cite{tevet2022human}.
`$\rightarrow$' indicates that closer alignment with the GT value is better.
}
\vspace{-2mm}
\centering
\setlength{\tabcolsep}{3pt} 
\scalebox{0.6}{
    \begin{tabular}{l c c c c c c c c c c c c c}
    \toprule
    \multirow{2}{*}{Method} & \multirow{2}{*}[-0.6ex]{\makecell[c]{Training-\\free}} & \multicolumn{4}{c}{First-half} & \multicolumn{4}{c}{Middle-half} & \multicolumn{4}{c}{Last-half} \\
    \cmidrule(lr){3-6} \cmidrule(lr){7-10} \cmidrule(lr){11-14}
     & & FID $\downarrow$   & Match. $\downarrow$  & R-prec. $\uparrow$ & Div. $\rightarrow$  & FID $\downarrow$   & Match. $\downarrow$  & R-prec. $\uparrow$ & Div. $\rightarrow$   & FID $\downarrow$   & Match. $\downarrow$  & R-prec. $\uparrow$ & Div. $\rightarrow$  \\
    \midrule
    CondMDI~\cite{cohan2024flexible} & N &  0.620 & 4.566 & 0.350 & 8.668 & \underline{0.594} & 4.489 & 0.354 & 8.618 &  0.362 & 4.409 & 0.365 & \textbf{9.050} \\
    Recon. Gui.~\cite{ho2022video} & Y &  11.342 & 5.230 & 0.284 & 6.101 & 12.806 & 5.152 & 0.299 & 6.041 &  7.322 & 4.866 & 0.319 & 6.804 \\
    Recon. Gui.*~\cite{ho2022video} & Y &  3.547 & 4.395 & 0.361 & 7.774 &  4.250 & 4.415 & 0.366 & 7.599 &  0.576 & \textbf{4.051} & 0.400 & 8.797 \\
    ALM (Ours) & Y &  \textbf{0.311} & \underline{4.143} & \underline{0.396} & \textbf{8.979} & \textbf{0.447} & \underline{4.175} & \textbf{0.395} & \textbf{8.820} &  \textbf{0.236} & 4.085 & \textbf{0.410} & \underline{9.041} \\
    ALM* (Ours)  & Y &  \underline{0.465} & \textbf{4.140} & \textbf{0.398} & \underline{8.828} & 0.645 & \textbf{4.159} & \underline{0.393} & \underline{8.695} &  \underline{0.260} & \underline{4.068} & \underline{0.408} & 9.025 \\
    \midrule
    Ground Truth & - & 0.001 & 3.243 & 0.453 & 9.299 & 0.001 & 3.243 & 0.453 & 9.299 & 0.001 & 3.243 & 0.453 & 9.299 \\
    \bottomrule
    \end{tabular}
    }
    \vspace{-6mm}
    \label{tab:motion_completion_cmp}
\end{table}

\subsection{3D Mesh Texturing}
\label{sec:exp_mesh}
\vspace{-2mm}
We extend our method to the 3D domain by applying it to the mesh texturing task. 
Following the standard setup of prior works~\cite{kim2024synctweedies, lee2025syncsde}, we sample 10 partially overlapping viewpoints around each mesh. 
Iterating over the viewpoints, we generate a rendered image from the current view using the pretrained depth-conditioned ControlNet~\cite{zhang2023adding}, where the overlapping regions of the pre-generated views are provided as additional conditioning.
After obtaining multi-view renderings, we bake them into a single texture map.

\begin{figure}[h!]
	\centering
	\includegraphics[width=1.0\linewidth]{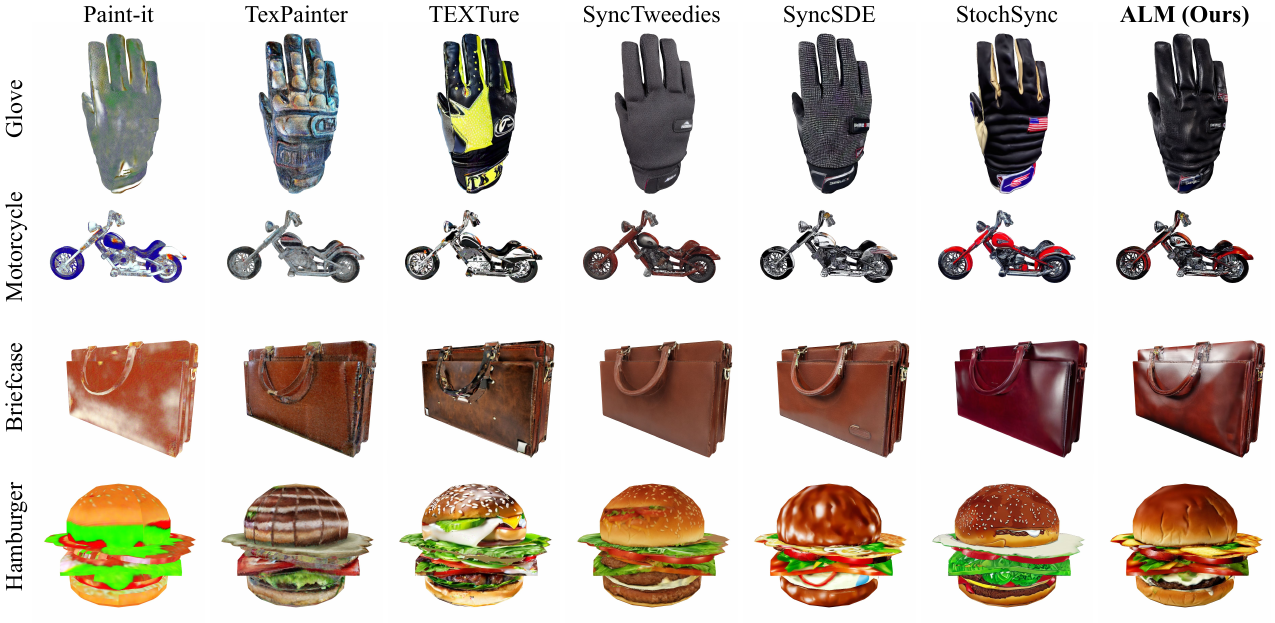}
	\vspace{-4mm}
	\caption{Qualitative comparison of ALM with baselines~\cite{youwang2024paint, zhang2024texpainter, richardson2023texture, kim2024synctweedies, lee2025syncsde, yeo2025stochsync} on 3D mesh texturing.
    ALM generates high-fidelity texture maps, outperforming prior works.}
	\vspace{-10mm}
\label{fig:mesh_texturing_cmp}
\end{figure}

\begin{figure}[h!]
	\centering
	\includegraphics[width=1.0\linewidth]{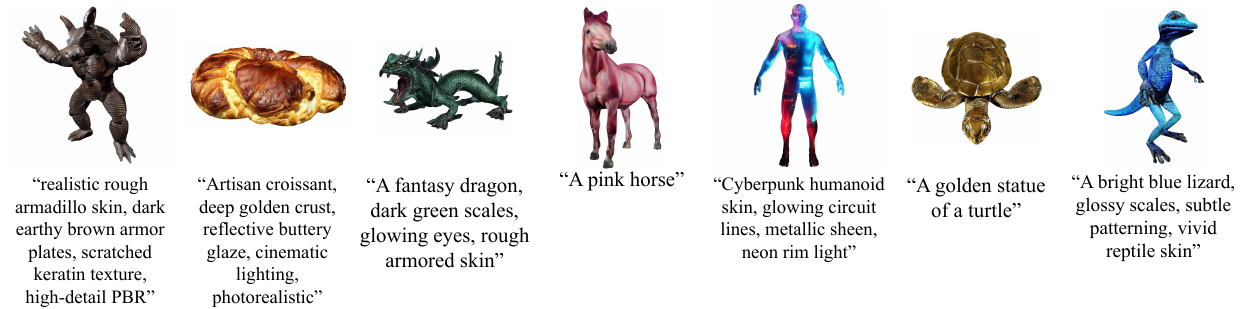}
	\vspace{-4mm}
	\caption{Additional qualitative results of 3D mesh texturing. ALM generates diverse and high-fidelity textures, effectively handling detailed prompts.}
\label{fig:mesh_texturing_ours}
\end{figure}

For quantitative evaluation, we use 250 mesh--prompt pairs sampled from the Objaverse dataset~\cite{objaverse}.
We compare our method against both task-specific methods~\cite{youwang2024paint, zhang2024texpainter, richardson2023texture} and synchronization-based approaches~\cite{kim2024synctweedies, lee2025syncsde, yeo2025stochsync}. 
Each textured mesh is rendered from 10 viewpoints, and the resulting images are used for evaluation.
We report widely adopted metrics including FID~\cite{heusel2017gans}, KID~\cite{binkowski2018demystifying}, and CLIP similarity~\cite{radford2021learning} between rendered images and text prompts. 
Reference images used to compute FID and KID are also generated using depth-conditioned ControlNet, based on the depth maps rendered from the same 10 viewpoints.
Results are presented in Table~\ref{tab:mesh_texturing_cmp} and Figure~\ref{fig:mesh_texturing_cmp}-\ref{fig:mesh_texturing_ours}. 
As shown, our method outperforms the baselines, further highlighting its effectiveness.

\vspace{-4mm}
\begin{table}[h!]
\centering
\caption{Quantitative evaluation on 3D mesh texturing. KID~\cite{binkowski2018demystifying} value is scaled by $10^3$.}
\vspace{-3mm}
\setlength{\tabcolsep}{4pt} 
\scalebox{0.65}{
\begin{tabular}{l c c c c c c c}
\toprule
{Method}  & {\makecell[c]{Paint-it~\cite{youwang2024paint}}} & {\makecell[c]{TexPainter~\cite{zhang2024texpainter}}} & {\makecell[c]{TEXTure~\cite{richardson2023texture}}} & {\makecell[c]{SyncTweedies~\cite{kim2024synctweedies}}} & {\makecell[c]{SyncSDE~\cite{lee2025syncsde}}} & {\makecell[c]{StochSync~\cite{yeo2025stochsync}}} & {\makecell[c]{ALM~(Ours)}}  \\
\midrule
FID $\downarrow$ &   200.89 & 191.17 & 184.66 & \underline{156.78} & 165.04 & 163.24 & \textbf{155.46}  \\
KID $\downarrow$ &    126.40 & 112.48 & 94.69 & 82.91 & \underline{82.75}    & 82.83 & \textbf{74.41} \\
CLIP-Sim. $\uparrow$ &    0.287 & 0.284 & 0.289 & \textbf{0.294} & 0.290 & 0.289 & \underline{0.292}  \\
\bottomrule
\end{tabular}
}
\vspace{-10mm}
\label{tab:mesh_texturing_cmp}
\end{table}

%


\section{Conclusion}
\label{sec:conclusion}

In this work, we introduce a novel, training-free sampling strategy for diffusion-based versatile content generation.
Such tasks go beyond simple generation by producing high-quality outputs conditioned on pre-generated inputs, encompassing a wide range of real-world applications.
Although diffusion models achieve remarkable performance in standard generation tasks, adapting them to such conditional settings typically requires expensive task-specific training, which limits generalizability.
To overcome this limitation, we propose a broadly applicable mechanism that transforms a wide range of pretrained generative models into a flexible content generation framework.
We synchronize pre-generated content with unobserved variables by maximizing a likelihood-based objective that combines a conditional likelihood term with a joint log-density term.
Furthermore, we introduce an acceleration strategy to improve computational efficiency.
Extensive experiments across diverse tasks and modalities demonstrate that our method achieves state-of-the-art performance.

\section*{Acknowledgements}

This work was supported by the BK21 FOUR program of the Education and Research Program for Future ICT Pioneers, Seoul National University in 2026, the National Research Foundation of Korea(NRF) grant (No. RS-2026-25485899) and Institute of Information \& communications Technology Planning \& Evaluation (IITP) grant [RS-2021-II211343, Artificial Intelligence Graduate School Program (Seoul National University)] funded by the Korea government(MSIT).

\bibliographystyle{splncs04}
\bibliography{main}

@String(IJCV  = {Int. J. Comput. Vis.})

@String(CVPR  = {IEEE Conf. Comput. Vis. Pattern Recog.})

@String(ICCV  = {Int. Conf. Comput. Vis.})

@String(ECCV  = {Eur. Conf. Comput. Vis.})

@String(NeurIPS = {Adv. Neural Inform. Process. Syst.})

@String(ICML  = {Int. Conf. Mach. Learn.})

@String(ICLR  = {Int. Conf. Learn. Represent.})

@String(CVPRW = {IEEE Conf. Comput. Vis. Pattern Recog. Worksh.})

@String(TOG   = {ACM Trans. Graph.})

@String(IJCV  = {IJCV})

@String(CVPR  = {CVPR})

@String(ICCV  = {ICCV})

@String(ECCV  = {ECCV})

@String(NeurIPS = {NeurIPS})

@String(ICML  = {ICML})

@String(ICLR  = {ICLR})

@String(CVPRW = {CVPRW})

@String(TOG   = {ACM TOG})

@inproceedings{ju2024brushnet,
  title={Brushnet: A plug-and-play image inpainting model with decomposed dual-branch diffusion},
  author={Ju, Xuan and Liu, Xian and Wang, Xintao and Bian, Yuxuan and Shan, Ying and Xu, Qiang},
  booktitle={ECCV},
  year={2024},
}

@inproceedings{zhuang2024task,
  title={A task is worth one word: Learning with task prompts for high-quality versatile image inpainting},
  author={Zhuang, Junhao and Zeng, Yanhong and Liu, Wenran and Yuan, Chun and Chen, Kai},
  booktitle={ECCV},
  year={2024},
}

@inproceedings{cohan2024flexible,
  title={Flexible motion in-betweening with diffusion models},
  author={Cohan, Setareh and Tevet, Guy and Reda, Daniele and Peng, Xue Bin and van de Panne, Michiel},
  booktitle={SIGGRAPH},
  year={2024}
}

@inproceedings{ronneberger2015u,
  title={U-net: Convolutional networks for biomedical image segmentation},
  author={Ronneberger, Olaf and Fischer, Philipp and Brox, Thomas},
  booktitle={International Conference on Medical image computing and computer-assisted intervention},
  year={2015},
}

@inproceedings{manukyan2023hd,
  title={Hd-painter: high-resolution and prompt-faithful text-guided image inpainting with diffusion models},
  author={Manukyan, Hayk and Sargsyan, Andranik and Atanyan, Barsegh and Wang, Zhangyang and Navasardyan, Shant and Shi, Humphrey},
  booktitle={ICLR},
  year={2025}
}

@incollection{perez2023poisson,
  title={Poisson image editing},
  author={P{\'e}rez, Patrick and Gangnet, Michel and Blake, Andrew},
  booktitle={Seminal Graphics Papers: Pushing the Boundaries, Volume 2},
  year={2023}
}

@article{avrahami2023blended,
  title={Blended latent diffusion},
  author={Avrahami, Omri and Fried, Ohad and Lischinski, Dani},
  journal={SIGGRAPH},
  year={2023},
}

@article{ho2022video,
  title={Video diffusion models},
  author={Ho, Jonathan and Salimans, Tim and Gritsenko, Alexey and Chan, William and Norouzi, Mohammad and Fleet, David J},
  journal={NeurIPS},
  year={2022}
}

@article{kim2024synctweedies,
  title={Synctweedies: A general generative framework based on synchronized diffusions},
  author={Kim, Jaihoon and Koo, Juil and Yeo, Kyeongmin and Sung, Minhyuk},
  journal={NeurIPS},
  year={2024}
}

@inproceedings{lee2025syncsde,
  title={SyncSDE: A Probabilistic Framework for Diffusion Synchronization},
  author={Lee, Hyunjun and Lee, Hyunsoo and Han, Sookwan},
  booktitle={CVPR},
  year={2025}
}

@article{ho2020denoising,
  title={Denoising diffusion probabilistic models},
  author={Ho, Jonathan and Jain, Ajay and Abbeel, Pieter},
  journal={NeurIPS},
  year={2020}
}

@article{song2020denoising,
  title={Denoising diffusion implicit models},
  author={Song, Jiaming and Meng, Chenlin and Ermon, Stefano},
  journal={ICLR},
  year={2021}
}

@article{song2020score,
  title={Score-based generative modeling through stochastic differential equations},
  author={Song, Yang and Sohl-Dickstein, Jascha and Kingma, Diederik P and Kumar, Abhishek and Ermon, Stefano and Poole, Ben},
  journal={ICLR},
  year={2021}
}

@inproceedings{rombach2022high,
  title={High-resolution image synthesis with latent diffusion models},
  author={Rombach, Robin and Blattmann, Andreas and Lorenz, Dominik and Esser, Patrick and Ommer, Bj{\"o}rn},
  booktitle={CVPR},
  year={2022}
}

@article{karras2022elucidating,
  title={Elucidating the design space of diffusion-based generative models},
  author={Karras, Tero and Aittala, Miika and Aila, Timo and Laine, Samuli},
  journal={NeurIPS},
  year={2022}
}

@article{kim2024fifo,
  title={Fifo-diffusion: Generating infinite videos from text without training},
  author={Kim, Jihwan and Kang, Junoh and Choi, Jinyoung and Han, Bohyung},
  journal={NeurIPS},
  year={2024}
}

@article{paszke2019pytorch,
  title={Pytorch: An imperative style, high-performance deep learning library},
  author={Paszke, Adam and Gross, Sam and Massa, Francisco and Lerer, Adam and Bradbury, James and Chanan, Gregory and Killeen, Trevor and Lin, Zeming and Gimelshein, Natalia and Antiga, Luca and others},
  journal={NeurIPS},
  year={2019}
}

@article{wang2025lavie,
  title={Lavie: High-quality video generation with cascaded latent diffusion models},
  author={Wang, Yaohui and Chen, Xinyuan and Ma, Xin and Zhou, Shangchen and Huang, Ziqi and Wang, Yi and Yang, Ceyuan and He, Yinan and Yu, Jiashuo and Yang, Peiqing and others},
  journal={IJCV},
  year={2025},
}

@article{ho2022classifier,
  title={Classifier-free diffusion guidance},
  author={Ho, Jonathan and Salimans, Tim},
  journal={NeurIPS Workshop},
  year={2021}
}

@article{karras2017progressive,
  title={Progressive growing of gans for improved quality, stability, and variation},
  author={Karras, Tero and Aila, Timo and Laine, Samuli and Lehtinen, Jaakko},
  journal={ICLR},
  year={2018}
}

@inproceedings{lugmayr2022repaint,
  title={Repaint: Inpainting using denoising diffusion probabilistic models},
  author={Lugmayr, Andreas and Danelljan, Martin and Romero, Andres and Yu, Fisher and Timofte, Radu and Van Gool, Luc},
  booktitle={CVPR},
  year={2022}
}

@inproceedings{choi2020stargan,
  title={Stargan v2: Diverse image synthesis for multiple domains},
  author={Choi, Yunjey and Uh, Youngjung and Yoo, Jaejun and Ha, Jung-Woo},
  booktitle={CVPR},
  year={2020}
}

@inproceedings{anoosheh2018combogan,
  title={Combogan: Unrestrained scalability for image domain translation},
  author={Anoosheh, Asha and Agustsson, Eirikur and Timofte, Radu and Van Gool, Luc},
  booktitle={CVPRW},
  year={2018}
}

@inproceedings{zhang2018perceptual,
  title={The Unreasonable Effectiveness of Deep Features as a Perceptual Metric},
  author={Zhang, Richard and Isola, Phillip and Efros, Alexei A and Shechtman, Eli and Wang, Oliver},
  booktitle={CVPR},
  year={2018}
}

@inproceedings{radford2021learning,
  title={Learning transferable visual models from natural language supervision},
  author={Radford, Alec and Kim, Jong Wook and Hallacy, Chris and Ramesh, Aditya and Goh, Gabriel and Agarwal, Sandhini and Sastry, Girish and Askell, Amanda and Mishkin, Pamela and Clark, Jack and others},
  booktitle={ICML},
  year={2021},
}

@article{kingma2013auto,
  title={Auto-encoding variational bayes},
  author={Kingma, Diederik P and Welling, Max},
  journal={arXiv:1312.6114},
  year={2013}
}

@article{heusel2017gans,
  title={Gans trained by a two time-scale update rule converge to a local nash equilibrium},
  author={Heusel, Martin and Ramsauer, Hubert and Unterthiner, Thomas and Nessler, Bernhard and Hochreiter, Sepp},
  journal={NIPS},
  year={2017}
}

@article{binkowski2018demystifying,
  title={Demystifying mmd gans},
  author={Bi{\'n}kowski, Miko{\l}aj and Sutherland, Danica J and Arbel, Michael and Gretton, Arthur},
  journal={ICLR},
  year={2018}
}

@article{schuhmann2022laion,
  title={Laion-5b: An open large-scale dataset for training next generation image-text models},
  author={Schuhmann, Christoph and Beaumont, Romain and Vencu, Richard and Gordon, Cade and Wightman, Ross and Cherti, Mehdi and Coombes, Theo and Katta, Aarush and Mullis, Clayton and Wortsman, Mitchell and others},
  journal={NeurIPS},
  year={2022}
}

@article{tevet2022human,
  title={Human motion diffusion model},
  author={Tevet, Guy and Raab, Sigal and Gordon, Brian and Shafir, Yonatan and Cohen-Or, Daniel and Bermano, Amit H},
  journal={ICLR},
  year={2023}
}

@inproceedings{guo2022generating,
  title={Generating diverse and natural 3d human motions from text},
  author={Guo, Chuan and Zou, Shihao and Zuo, Xinxin and Wang, Sen and Ji, Wei and Li, Xingyu and Cheng, Li},
  booktitle={CVPR},
  year={2022}
}

@inproceedings{corneanu2024latentpaint,
  title={Latentpaint: Image inpainting in latent space with diffusion models},
  author={Corneanu, Ciprian and Gadde, Raghudeep and Martinez, Aleix M},
  booktitle={WACV},
  year={2024}
}

@inproceedings{bar2023multidiffusion,
  title={Multidiffusion: Fusing diffusion paths for controlled image generation},
  author={Bar-Tal, Omer and Yariv, Lior and Lipman, Yaron and Dekel, Tali},
  booktitle={ICML},
  year={2023}
}

@article{qiu2023freenoise,
  title={Freenoise: Tuning-free longer video diffusion via noise rescheduling},
  author={Qiu, Haonan and Xia, Menghan and Zhang, Yong and He, Yingqing and Wang, Xintao and Shan, Ying and Liu, Ziwei},
  journal={ICLR},
  year={2024}
}

@inproceedings{chen2023seine,
  title={Seine: Short-to-long video diffusion model for generative transition and prediction},
  author={Chen, Xinyuan and Wang, Yaohui and Zhang, Lingjun and Zhuang, Shaobin and Ma, Xin and Yu, Jiashuo and Wang, Yali and Lin, Dahua and Qiao, Yu and Liu, Ziwei},
  booktitle={ICLR},
  year={2024}
}

@inproceedings{karunratanakul2023guided,
  title={Guided motion diffusion for controllable human motion synthesis},
  author={Karunratanakul, Korrawe and Preechakul, Konpat and Suwajanakorn, Supasorn and Tang, Siyu},
  booktitle={ICCV},
  year={2023}
}

@article{podell2023sdxl,
  title={Sdxl: Improving latent diffusion models for high-resolution image synthesis},
  author={Podell, Dustin and English, Zion and Lacey, Kyle and Blattmann, Andreas and Dockhorn, Tim and M{\"u}ller, Jonas and Penna, Joe and Rombach, Robin},
  journal={ICLR},
  year={2024}
}

@article{unterthiner2019fvd,
  title={FVD: A new metric for video generation},
  author={Unterthiner, Thomas and Van Steenkiste, Sjoerd and Kurach, Karol and Marinier, Rapha{\"e}l and Michalski, Marcin and Gelly, Sylvain},
  booktitle={ICLR Workshop},
  year={2019}
}

@article{yeo2025stochsync,
  title={Stochsync: Stochastic diffusion synchronization for image generation in arbitrary spaces},
  author={Yeo, Kyeongmin and Kim, Jaihoon and Sung, Minhyuk},
  journal={ICLR},
  year={2025}
}

@article{yan2025flexpainter,
  title={FlexPainter: Flexible and Multi-View Consistent Texture Generation},
  author={Yan, Dongyu and Wu, Leyi and Lin, Jiantao and Wang, Luozhou and Xu, Tianshuo and Chen, Zhifei and Yang, Zhen and Xu, Lie and Zhang, Shunsi and Chen, Yingcong},
  journal={arXiv:2506.02620},
  year={2025}
}

@inproceedings{youwang2024paint,
  title={Paint-it: Text-to-texture synthesis via deep convolutional texture map optimization and physically-based rendering},
  author={Youwang, Kim and Oh, Tae-Hyun and Pons-Moll, Gerard},
  booktitle={CVPR},
  year={2024}
}

@inproceedings{zhang2024texpainter,
  title={Texpainter: Generative mesh texturing with multi-view consistency},
  author={Zhang, Hongkun and Pan, Zherong and Zhang, Congyi and Zhu, Lifeng and Gao, Xifeng},
  booktitle={SIGGRAPH},
  year={2024}
}

@inproceedings{richardson2023texture,
  title={Texture: Text-guided texturing of 3d shapes},
  author={Richardson, Elad and Metzer, Gal and Alaluf, Yuval and Giryes, Raja and Cohen-Or, Daniel},
  booktitle={SIGGRAPH},
  year={2023}
}

@misc{flux2024,
    author={Black Forest Labs},
    title={FLUX},
    year={2024},
    howpublished={\url{https://github.com/black-forest-labs/flux}},
    note={Accessed: June 29, 2026}
}

@article{zhang2011fsim,
  title={FSIM: A feature similarity index for image quality assessment},
  author={Zhang, Lin and Zhang, Lei and Mou, Xuanqin and Zhang, David},
  journal={IEEE transactions on Image Processing},
  year={2011}
}

@article{wu2023q,
  title={Q-align: Teaching lmms for visual scoring via discrete text-defined levels},
  author={Wu, Haoning and Zhang, Zicheng and Zhang, Weixia and Chen, Chaofeng and Liao, Liang and Li, Chunyi and Gao, Yixuan and Wang, Annan and Zhang, Erli and Sun, Wenxiu and others},
  journal={ICML},
  year={2024}
}

@inproceedings{esser2024scaling,
  title={Scaling rectified flow transformers for high-resolution image synthesis},
  author={Esser, Patrick and Kulal, Sumith and Blattmann, Andreas and Entezari, Rahim and M{\"u}ller, Jonas and Saini, Harry and Levi, Yam and Lorenz, Dominik and Sauer, Axel and Boesel, Frederic and others},
  booktitle={ICML},
  year={2024}
}

@misc{wu2025qwenimagetechnicalreport,
      title={Qwen-Image Technical Report}, 
      author={Chenfei Wu and Jiahao Li and Jingren Zhou and Junyang Lin and Kaiyuan Gao and Kun Yan and Sheng-ming Yin and Shuai Bai and Xiao Xu and Yilei Chen and Yuxiang Chen and Zecheng Tang and Zekai Zhang and Zhengyi Wang and An Yang and Bowen Yu and Chen Cheng and Dayiheng Liu and Deqing Li and Hang Zhang and Hao Meng and Hu Wei and Jingyuan Ni and Kai Chen and Kuan Cao and Liang Peng and Lin Qu and Minggang Wu and Peng Wang and Shuting Yu and Tingkun Wen and Wensen Feng and Xiaoxiao Xu and Yi Wang and Yichang Zhang and Yongqiang Zhu and Yujia Wu and Yuxuan Cai and Zenan Liu},
      year={2025},
      eprint={2508.02324},
      archivePrefix={arXiv},
      primaryClass={cs.CV},
      url={https://arxiv.org/abs/2508.02324}, 
      note={Accessed: June 29, 2026}
}

@inproceedings{peebles2023scalable,
  title={Scalable diffusion models with transformers},
  author={Peebles, William and Xie, Saining},
  booktitle={ICCV},
  year={2023}
}

@inproceedings{zhang2023adding,
  title={Adding conditional control to text-to-image diffusion models},
  author={Zhang, Lvmin and Rao, Anyi and Agrawala, Maneesh},
  booktitle={ICCV},
  year={2023}
}

@article{objaverse,
  title={Objaverse: A Universe of Annotated 3D Objects},
  author={Matt Deitke and Dustin Schwenk and Jordi Salvador and Luca Weihs and Oscar Michel and Eli VanderBilt and Ludwig Schmidt and Kiana Ehsani and Aniruddha Kembhavi and Ali Farhadi},
  journal={CVPR},
  year={2023}
}

@article{lipman2022flow,
  title={Flow matching for generative modeling},
  author={Lipman, Yaron and Chen, Ricky TQ and Ben-Hamu, Heli and Nickel, Maximilian and Le, Matt},
  journal={ICLR},
  year={2023}
}

@article{liu2022flow,
  title={Flow straight and fast: Learning to generate and transfer data with rectified flow},
  author={Liu, Xingchao and Gong, Chengyue and Liu, Qiang},
  journal={ICLR},
  year={2023}
}

@article{saharia2022photorealistic,
  title={Photorealistic text-to-image diffusion models with deep language understanding},
  author={Saharia, Chitwan and Chan, William and Saxena, Saurabh and Li, Lala and Whang, Jay and Denton, Emily L and Ghasemipour, Kamyar and Gontijo Lopes, Raphael and Karagol Ayan, Burcu and Salimans, Tim and others},
  journal={NeurIPS},
  year={2022}
}

@article{HaCohen2024LTXVideo,
  title={LTX-Video: Realtime Video Latent Diffusion},
  author={HaCohen, Yoav and Chiprut, Nisan and Brazowski, Benny and Shalem, Daniel and Moshe, Dudu and Richardson, Eitan and Levin, Eran and Shiran, Guy and Zabari, Nir and Gordon, Ori and Panet, Poriya and Weissbuch, Sapir and Kulikov, Victor and Bitterman, Yaki and Melumian, Zeev and Bibi, Ofir},
  journal={arXiv:2501.00103},
  year={2024}
}

@inproceedings{
  yang2024cogvideox,
  title={CogVideoX: Text-to-Video Diffusion Models with An Expert Transformer},
  author={Zhuoyi Yang and Jiayan Teng and Wendi Zheng and Ming Ding and Shiyu Huang and Jiazheng Xu and Yuanming Yang and Wenyi Hong and Xiaohan Zhang and Guanyu Feng and Da Yin and Yuxuan.Zhang and Weihan Wang and Yean Cheng and Bin Xu and Xiaotao Gu and Yuxiao Dong and Jie Tang},
  booktitle={ICLR},
  year={2025}
}

@article{hong2022cogvideo,
  title={CogVideo: Large-scale Pretraining for Text-to-Video Generation via Transformers},
  author={Hong, Wenyi and Ding, Ming and Zheng, Wendi and Liu, Xinghan and Tang, Jie},
  journal={ICLR},
  year={2023}
}

@misc{kakaobrain2022coyo,
  title         = {COYO-700M: Image-Text Pair Dataset},
  author        = {Byeon, Minwoo and Park, Beomhee and Kim, Haecheon and Lee, Sungjun and Baek, Woonhyuk and Kim, Saehoon},
  year          = {2022},
  howpublished  = {\url{https://github.com/kakaobrain/coyo-dataset}},
  note          = {Accessed: June 29, 2026}
}

@article{lee2023syncdiffusion,
  title={Syncdiffusion: Coherent montage via synchronized joint diffusions},
  author={Lee, Yuseung and Kim, Kunho and Kim, Hyunjin and Sung, Minhyuk},
  journal={NeurIPS},
  year={2023}
}

@inproceedings{zeng2024paint3d,
  title={Paint3d: Paint anything 3d with lighting-less texture diffusion models},
  author={Zeng, Xianfang and Chen, Xin and Qi, Zhongqi and Liu, Wen and Zhao, Zibo and Wang, Zhibin and Fu, Bin and Liu, Yong and Yu, Gang},
  booktitle={CVPR},
  year={2024}
}

@inproceedings{bain2021frozen,
  title={Frozen in time: A joint video and image encoder for end-to-end retrieval},
  author={Bain, Max and Nagrani, Arsha and Varol, G{\"u}l and Zisserman, Andrew},
  booktitle={ICCV},
  year={2021}
}

@inproceedings{sohl2015deep,
  title={Deep unsupervised learning using nonequilibrium thermodynamics},
  author={Sohl-Dickstein, Jascha and Weiss, Eric and Maheswaranathan, Niru and Ganguli, Surya},
  booktitle={ICML},
  year={2015}
}

@article{lee2023conditional,
  title={Conditional score guidance for text-driven image-to-image translation},
  author={Lee, Hyunsoo and Kang, Minsoo and Han, Bohyung},
  journal={NeurIPS},
  year={2023}
}

@inproceedings{liu2024text,
  title={Text-guided texturing by synchronized multi-view diffusion},
  author={Liu, Yuxin and Xie, Minshan and Liu, Hanyuan and Wong, Tien-Tsin},
  booktitle={SIGGRAPH Asia},
  year={2024}
}

@inproceedings{wang2003multiscale,
  title={Multiscale structural similarity for image quality assessment},
  author={Wang, Zhou and Simoncelli, Eero P and Bovik, Alan C},
  booktitle={The thrity-seventh asilomar conference on signals, systems \& computers, 2003},
  year={2003}
}

@article{aurenhammer1991voronoi,
  title={Voronoi diagrams—a survey of a fundamental geometric data structure},
  author={Aurenhammer, Franz},
  journal={ACM computing surveys (CSUR)},
  year={1991},
}

@article{ravi2020pytorch3d,
    author = {Nikhila Ravi and Jeremy Reizenstein and David Novotny and Taylor Gordon
                  and Wan-Yen Lo and Justin Johnson and Georgia Gkioxari},
    title = {Accelerating 3D Deep Learning with PyTorch3D},
    journal = {arXiv:2007.08501},
    year = {2020},
}

@article{Laine2020diffrast,
  title   = {Modular Primitives for High-Performance Differentiable Rendering},
  author  = {Samuli Laine and Janne Hellsten and Tero Karras and Yeongho Seol and Jaakko Lehtinen and Timo Aila},
  journal = {ACM TOG},
  year    = {2020}
}

@article{liang2025UnitTEX,
  title={UniTEX: Universal High Fidelity Generative Texturing for 3D Shapes},
  author={Yixun Liang and Kunming Luo and Xiao Chen and Rui Chen and Hongyu Yan and Weiyu Li and Jiarui Liu and Ping Tan},
  journal={CVPR},
  year={2026}
}

@article{chung2022diffusion,
  title={Diffusion posterior sampling for general noisy inverse problems},
  author={Chung, Hyungjin and Kim, Jeongsol and Mccann, Michael T and Klasky, Marc L and Ye, Jong Chul},
  journal={ICLR},
  year={2023}
}

@article{pandey2025variational,
  title={Variational control for guidance in diffusion models},
  author={Pandey, Kushagra and Sofian, Farrin Marouf and Draxler, Felix and Karaletsos, Theofanis and Mandt, Stephan},
  journal={ICML},
  year={2025}
}

@article{geyfman2026calibrated,
  title={Calibrated test-time guidance for bayesian inference},
  author={Geyfman, Daniel and Draxler, Felix and Groeneveld, Jan and Lee, Hyunsoo and Karaletsos, Theofanis and Mandt, Stephan},
  journal={ICML},
  year={2026}
}

@inproceedings{song2023pseudoinverse,
  title={Pseudoinverse-guided diffusion models for inverse problems},
  author={Song, Jiaming and Vahdat, Arash and Mardani, Morteza and Kautz, Jan},
  booktitle={ICLR},
  year={2023}
}

@InProceedings{Liu_2026_CVPR,
    author    = {Liu, Yunpeng and Hou, Xingzhong and Wu, Jie and Liu, Boxiao and Zhang, Yi and Song, Guanglu and Liu, Yu and Tian, Changyao and Luo, Gen and You, Haihang},
    title     = {Blend-Aware Latent Diffusion: Mitigating Stitched Seams in Image Inpainting},
    booktitle = {CVPR Findings},
    year      = {2026}
}

@InProceedings{Lai_2026_CVPR,
    author    = {Lai, Zeqiang and Zhao, Yunfei and Zhao, Zibo and Yang, Xin and Huang, Xin and Huang, Jingwei and Yue, Xiangyu and Guo, Chunchao},
    title     = {NaTex: Seamless Texture Generation as Latent Color Diffusion},
    booktitle = {CVPR},
    year      = {2026}
}

@InProceedings{Wu_2026_CVPR,
    author    = {Wu, Linjun and Yu, Jiejia and Jin, Leyang and Wang, He and Zheng, Bowen and Yang, Xu and Jiang, Hao and Xia, Fei and Ling, Fei and Deng, Jun and Jin, Xiaogang},
    title     = {Unifying Precise Keyframes and Semantic Control via Multi-level Diffusion},
    booktitle = {CVPR},
    year      = {2026}
}

\clearpage

\section*{Supplementary Material}


\setcounter{equation}{12}
\setcounter{table}{6}
\setcounter{figure}{8}

\appendix
\renewcommand{\theHsection}{appendix.\arabic{section}}

In this supplementary material, we first provide a detailed derivation of our method in Appendix~\ref{sec:supp_method_derivation}.
We then present additional experimental details and results in Appendix~\ref{sec:supp_additional_exp}, including image inpainting, wide image generation, human motion completion, 3D mesh texturing, and long video generation.
In Appendix~\ref{sec:supp_score_estimation}, we provide a justification for the joint score estimation used in the method derivation.
Appendix~\ref{sec:supp_acceleration} validates the practical applicability of the proposed acceleration strategy.
We also discuss the computational cost in Appendix~\ref{sec:supp_computation}, and show that our method is robust to the hyperparameter choices in Appendix~\ref{sec:supp_hyp}.

\section{Detailed Method Derivation}
\label{sec:supp_method_derivation}

\subsection{Unobserved Region Optimization}
\label{sec:supp_method_alm}

We aim to optimize the unobserved region of $\by_t$ by imposing a novel sampling strategy.
At each diffusion timestep $t$, we introduce an additional term $\dby_t$, which is added to $\by_t$.
We design $\dby_t= \sum_{i=1}^N {\dby_t^i}$, where the sequence $\{ \dby_t^i \}_{i=1}^N$ is constructed to iteratively minimize the following terms:
\begin{equation}
     -\lambda_1  \log  p(\bx_t , \bmask \mid \by_t^i + \bmask \odot \dby_t^i, \mathbf{c}) - \lambda_2  \log  p(\bx_t , \bmask , \by_t^i + \bmask \odot \dby_t^i \mid \mathbf{c}),
    \label{eq:supp_f_dby}
\end{equation}
with $\lambda_1$ and $\lambda_2$ being scalar hyperparameters ($\lambda_1 > \lambda_2$).
Note that $\by_t^i = \by_t^{i-1} + \bmask \odot \dby_t^{i-1}$, and the initial values are set as $\by_t^1= \by_t$ and $\{\dby_t^i\}_{i=1}^N = \{  \mathbf{0}\}_{i=1}^N$.

We distinguish between the roles of the conditional likelihood and joint log-density terms presented in Eq.~\eqref{eq:supp_f_dby}.
The conditional likelihood encourages contextual consistency by aligning the unobserved region with the pre-generated content, whereas the joint log-density term encourages the blended content to lie within high-density regions of the full data distribution, thereby harmonizing both regions into a globally realistic sample.
As in the main text, this corresponds to likelihood maximization in a score-based sense, since the update follows score estimates of the corresponding composite log-probability objective over the unobserved variable while keeping the pretrained model fixed.
This separation enables our method to simultaneously preserve local consistency and improve global harmonization.
We verify that using both terms is essential for high-quality content generation in Sec. 4.2.
The coefficients $\lambda_1$ and $\lambda_2$ act as weights in a composite energy function~\cite{song2020score}, allowing adaptive balancing between two terms for better performance.

We define $f(\dby_t^i)$ as the optimization objective defined in Eq.~\eqref{eq:supp_f_dby}.
By assuming $\| \dby_t^i \| \ll 1$ and applying a Taylor expansion around $\mathbf{0}$, we derive:
\begin{align}
	f(&\dby_t^i) \simeq  -\lambda_1 \left[	\log p(\bx_t, \bmask \mid \by_t^i, \mathbf{c}) + (\bmask \odot \nabla_{\by_t^i} \log p(\bx_t, \bmask \mid \by_t^i, \mathbf{c}))^{\top} \dby_t^i \right] \nonumber \\
	& -\lambda_2 \left[ \log p(\bx_t, \bmask, \by_t^i \mid \mathbf{c}) + (\bmask \odot \nabla_{\by_t^i}\log p(\bx_t, \bmask, \by_t^i \mid \mathbf{c}))^{\top} \dby_t^i \right].
\end{align}
Taking a gradient descent step with respect to $\dby_t$ with step size set to 1, we obtain:
\begin{equation}
    \dby_t^i =\bmask \odot (\lambda_1 \nabla_{\by_t^i} \log p(\bx_t, \bmask \mid \by_t^i, \mathbf{c}) + \lambda_2 \nabla_{\by_t^i} \log p(\bx_t, \bmask , \by_t^i \mid \mathbf{c})).
\end{equation}
Note that the small-magnitude constraint can be satisfied by choosing sufficiently small values of $\lambda_1$ and $\lambda_2$, which we detail in Appendix D.

Using Bayes' rule, we further factorize the conditional log-likelihood as follows:
\begin{equation}
    \nabla_{\by_t^i} \log p(\bx_t ,\bmask \mid \by_t^i, \mathbf{c}) = \nabla_{\by_t^i}  \log p(\bx_t, \bmask , \by_t^i \mid \mathbf{c}) - \nabla_{\by_t^i} \log  p(\by_t^i \mid \mathbf{c}).
\end{equation}
Following the score-based substitution technique~\cite{song2020score, lee2023conditional}, the second term is calculated using the pretrained diffusion model:
\begin{equation}
    \nabla_{\by_t^i} \log  p(\by_t^i \mid \mathbf{c}) \simeq - \frac{1}{\sqrt{1-\alpha_t}} \epsilon_{\theta}(\by_t^i, t, \mathbf{c})
    \label{eq:supp_grad_log_y}
\end{equation}
For the first term, we define $\nabla_{\by_t^i} \log p(\bx_t, \bmask, \by_t^i \mid \mathbf{c}) \simeq \nabla_{\by_t^i} \log p(\be_t^i \mid \mathbf{c})$.
We justify that this score estimation works well in Appendix C.
Then we get
\begin{equation}
    \nabla_{\by_t^i} \log p(\be_t^i \mid \mathbf{c}) = \nabla_{\be_t^i} \log p(\be_t^i \mid \mathbf{c}) \odot \bmask \simeq - \frac{1}{\sqrt{1-\alpha_t}}\epsilon_{\theta} (\be_t^i, t, \mathbf{c}) \odot \bmask .
    \label{eq:supp_grad_log_e}
\end{equation}
Putting these together, the closed form formula for $\dby_t$ becomes
\begin{equation}
    \dby_t^i = \bmask \odot (\lambda_1 (\epsilon_{\theta}({\by}_t^i, t, \mathbf{c}) - \epsilon_{\theta}(\be_t^i ,t, \mathbf{c})) - \lambda_2 \epsilon_{\theta}(\be_t^i, t, \mathbf{c})),
    \label{eq:supp_dby_closed_form}
\end{equation}
up to a scaling factor of $1/\sqrt{1-\alpha_t}$.

\subsection{Acceleration strategy}

From the relation of 
$ 
    \|\dby_t^i \| \leq | \lambda_1| \| \epsilon_{\theta}({\by}_t^i, t, \mathbf{c}) - \epsilon_{\theta}(\be_t^i, t, \mathbf{c}) \| + | \lambda_2 | \| \epsilon_{\theta}(\be_t^i, t, \mathbf{c}) \|,
$
we can choose $\lambda_1$ and $\lambda_2$ such that each $\dby_t^i$ remains sufficiently small for accurate Taylor expansion, then get a sequence $\{ \dby_t^i \}_{i=1}^N$ with $N$ iterations.
However, this iterative process is computationally expensive, since its time complexity scales as $\mathcal{O}(N)$.
To address this, we propose a \textit{one-step approximation} strategy, which collapses the effect of multiple small updates into a single large step without degrading the performance.
For the rest of the derivation, we denote $\by_t^i=\by_t^{i-1} + \dby_t^{i-1}$ from the definition of $\dby_t^{i-1}$.
Here, we present two claims for derivation:
\vspace{1mm}
\noindent \fbox{
\parbox{0.96\linewidth}{
\textbf{Claim 1.} $\dby_t^i$ is small enough for all $1 \leq i \leq N$. 
That is, $\lambda_1$ and $\lambda_2$ are chosen such that 
$\| \dby_t^i \| \ll 1$. \\
\textbf{Claim 2.} The noise prediction network $\epsilon_{\theta}(\cdot, \cdot, \cdot)$ of the pretrained diffusion model is $L$-Lipschitz~\cite{karras2022elucidating,kim2024fifo}.
}
}
\newline

\noindent Using these claims, we analyze the difference between $\dby_t^i$ and $\dby_t^{i+1}$:
\begin{align}
	\|\dby^{i+1}_t & - \dby_t^{i} \| \nonumber
	\\  \leq  & \; \| (\lambda_1 (\epsilon_{\theta}(\by_t^i + \Delta \by_t^i, t, \mathbf{c} ) - \epsilon_{\theta}(\be_t^i  + \dby_t^i, t, \mathbf{c}) )   - \lambda_2 \epsilon_{\theta}(\be_t^i+ \dby_t^i, t, \mathbf{c})  )  \nonumber \\
	& - (\lambda_1 (\epsilon_{\theta}(\by_t^i, t, \mathbf{c}) - \epsilon_{\theta}(\be_t^i, t, \mathbf{c})) - \lambda_2 \epsilon_{\theta}(\be_t^i, t, \mathbf{c})) \| \nonumber \\
	= & \; \| \lambda_1 (\epsilon_{\theta}(\by_t^i + \dby_t^i, t, \mathbf{c} ) - \epsilon_{\theta}(\by_t^i, t, \mathbf{c}) ) \nonumber \\
	& - (\lambda_1 + \lambda_2) (\epsilon_{\theta}(\be_t^i + \dby_t^i, t, \mathbf{c} ) - \epsilon_{\theta}(\be_t^i, t, \mathbf{c}) ) \| \nonumber \\
	\leq &\; \lambda_1 \| \epsilon_{\theta}(\by_t^i + \dby_t^i , t, \mathbf{c}) - \epsilon_{\theta}(\by_t^i, t, \mathbf{c}) \| \nonumber \\ 
	& + (\lambda_1 + \lambda_2 ) \| \epsilon_{\theta}(\be_t^i + \dby_t^i, t, \mathbf{c} ) - \epsilon_{\theta}(\be_t^i, t, \mathbf{c}) \| \nonumber \\
	\leq & \; L(2 \lambda_1 + \lambda_2) \| \Delta \by_t^i \|  = \mathcal{O}( \| \Delta \by_t^i \|)
\end{align}
From Claim 1, it follows that $\dby_t^{i+1} \simeq \dby_t^i$ for all $i$.
Therefore, we approximate the iterative update with a 1-step approximation as follows:
\begin{align}
	\dby_t \simeq N \dby^1_t & = \bmask \odot (N \lambda_1 (\epsilon_{\theta}(\by_t, t, \mathbf{c}) - \epsilon_{\theta}(\be_t, t, \mathbf{c})) - N \lambda_2 \epsilon_{\theta}(\be_t, t, \mathbf{c})) \nonumber  \\ 
	& = \bmask \odot (w_1' (\epsilon_{\theta}(\by_t, t, \mathbf{c}) - \epsilon_{\theta}(\be_t, t, \mathbf{c})) -  w_2 \epsilon_{\theta}(\be_t, t, \mathbf{c})),
\end{align}
where we define $w_1' = N \lambda_1$ and $w_2 = N \lambda_2$.
In practice, we set $w_1 = w_1'$, yielding only two hyperparameters.

We justify that when Claim 1 holds, $\| \dby_t^{i+1} - \dby_t^{i} \| \simeq 0$, making the one-step approximation valid in {Appendix D}. 
Note that the use of $\mathcal{O}(\cdot)$ bounds for analyzing diffusion dynamics is not uncommon in the literature~\cite{kim2024fifo}, supporting the reasonableness of our derivation with strong empirical results.
Thanks to this approximation, instead of gradually refining the unobserved region through $N$ iterations, we directly compute the outcome of the full optimization in a single update.
This technique significantly reduces computation time \textit{without sacrificing the performance} as verified in Sec. 4.2.
In practice, we apply a decaying schedule to hyperparameters to better ensure the small-update assumption, defined as:
\begin{equation}
    w_i = \sigma_t w_i^{\mathrm{init}}, \quad \sigma_t = \sqrt{\frac{1-\alpha_{t-1}}{1-\alpha_t}} \sqrt{1 - \frac{\alpha_t}{\alpha_{t-1}}}
\end{equation}
where $\sigma_t$ follows the same definition as in DDPM~\cite{ho2020denoising}.

\section{Additional Experimental Results}
\label{sec:supp_additional_exp}
In this section, we present additional versatile content generation results with task-specific experimental details.

\subsection{Image Inpainting}
\label{sec:supp_additional_inpainting}

\begin{figure}[h!]
	\centering
	\includegraphics[width=1.0\linewidth]{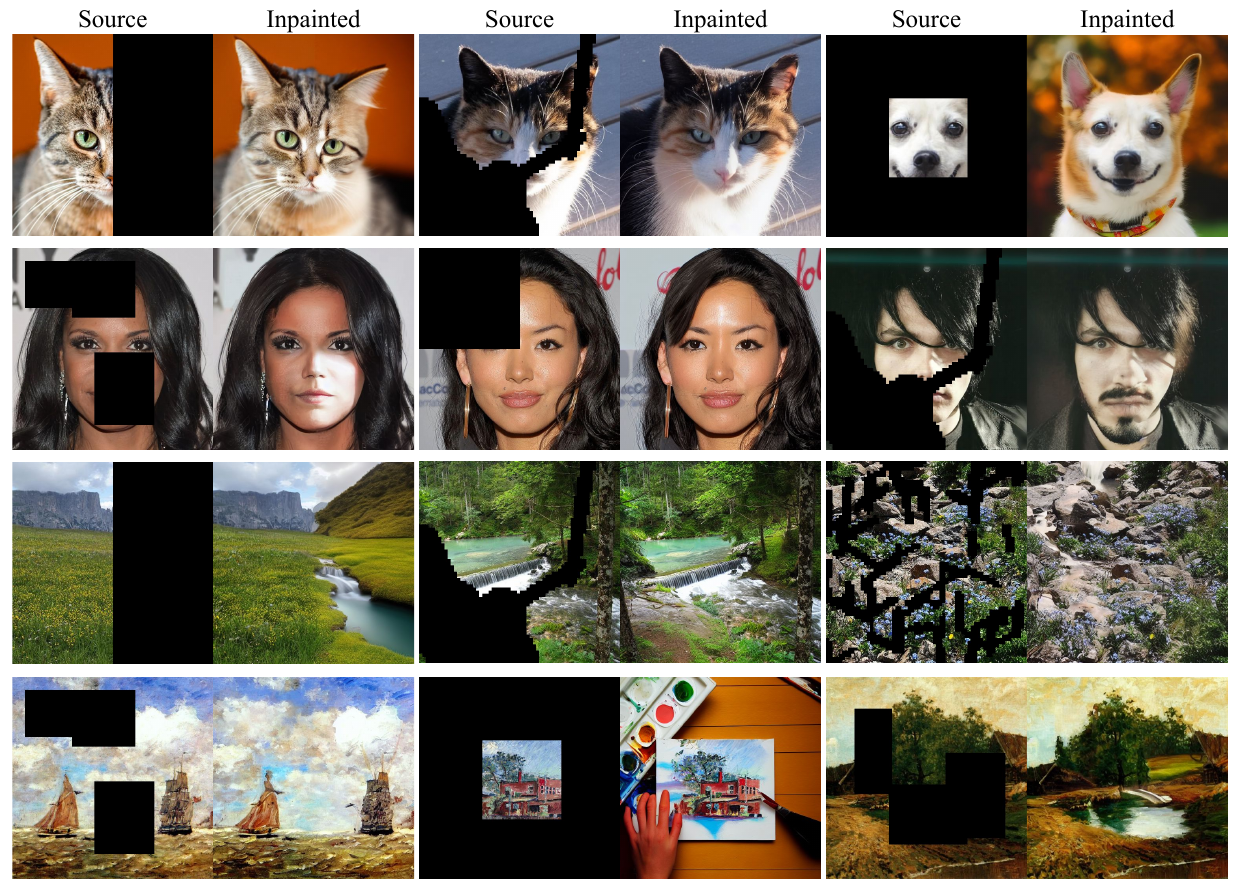}
	\vspace{-6mm}
	\caption{Additional qualitative results of image inpainting using the pretrained Stable Diffusion~\cite{rombach2022high} on diverse datasets~\cite{choi2020stargan, karras2017progressive, anoosheh2018combogan}.}
\label{fig:image_inpainting_supple}
\end{figure}

\paragraph{Additional evaluation using Stable Diffusion.}
We use the pretrained Stable Diffusion~\cite{rombach2022high} v1.5 model for the experiments, with 50 DDIM~\cite{song2020denoising} steps.
Additional evaluations on the image inpainting task are then conducted using the Seasons and Painters datasets~\cite{anoosheh2018combogan}. 
Following the setup described in Section~4.2, 1,000 image--mask pairs are sampled from each dataset.
Quantitative results are reported in Table~\ref{tab:image_inpainting_cmp_supp}.
For the blending operation (denoted as *), we follow the setting of  BrushNet~\cite{ju2024brushnet}, where the binary mask is first blurred with a Gaussian filter before blending.
Additional qualitative results obtained with the pretrained Stable Diffusion are shown in Figure~\ref{fig:image_inpainting_supple}.
As shown, our method achieves superior inpainting performance across diverse scenarios.

\vspace{-5mm}
\begin{table}[h!]
\centering
\caption{Additional quantitative evaluation on image inpainting. Methods with * and $^\dagger$ denote results obtained with pixel-level blending and super-resolution, respectively.}
\vspace{-4mm}
\setlength{\tabcolsep}{3pt} 
\scalebox{0.63}{
\begin{tabular}{l c c c c c c c c c c c}
\toprule
\multirow{2}{*}{Method} & \multirow{2}{*}[-0.6ex]{\makecell[c]{Training-\\free}} & \multicolumn{5}{c}{Seasons~\cite{anoosheh2018combogan}} & \multicolumn{5}{c}{Painters~\cite{anoosheh2018combogan}} \\
\cmidrule(lr){3-7} \cmidrule(lr){8-12}
 & & LPIPS $\downarrow$ & MSE $\downarrow$ & M-SSIM $\uparrow$ & MS-SSIM $\uparrow$ & FSIM $\uparrow$ & LPIPS $\downarrow$ & MSE $\downarrow$ & M-SSIM $\uparrow$ & MS-SSIM $\uparrow$ & FSIM $\uparrow$   \\
\midrule
BrushNet~\cite{ju2024brushnet}     & N &   0.337 & 
0.229 & 
0.200 & 
0.537 & 
0.708 &  0.351 & 
0.237 & 
0.157 & 
0.515 & 
0.699
 \\
PowerPaint~\cite{zhuang2024task}   & N &   0.345 & 
0.249 & 
0.210 & 
0.542 & 
0.704 &  0.365 & 
0.247 & 
0.164 & 
0.518 & 
0.690
 \\
SDI~\cite{rombach2022high}          & N &   \textbf{0.298} & 
\textbf{0.155} & 
\underline{0.245} & 
0.577 & 
\underline{0.731} &  \textbf{0.322} & 
\underline{0.163} & 
0.191 & 
\underline{0.558} & 
\underline{0.724}
 \\
SyncSDE~\cite{lee2025syncsde}      & Y &   0.350 & 
0.197 & 
0.227 & 
\underline{0.579} & 
0.725 &  0.395 & 
0.194 & 
\underline{0.195} & 
0.554 & 
0.707
 \\
HD-Painter~\cite{manukyan2023hd}   & Y & \underline{0.316} & 
0.163 & 
0.243 & 
0.569 & 
0.713 &  0.355 & 
0.200 & 
0.178 & 
0.543 & 
0.694
 \\
ALM (Ours)   & Y &   0.323 & 
\underline{0.162} & 
\textbf{0.278} & 
\textbf{0.626} & 
\textbf{0.748} &  \underline{0.351} & 
\textbf{0.153} & 
\textbf{0.236} & 
\textbf{0.605} & 
\textbf{0.729}
 \\
\midrule
BrushNet*~\cite{ju2024brushnet}    & N &  0.291 & 
0.202 & 
0.228 & 
0.602 & 
\underline{0.754} &  \underline{0.302} & 
0.216 & 
0.182 & 
0.579 & 
\underline{0.745}
 \\
HD-Painter*$^\dagger$~\cite{manukyan2023hd} & Y &  \underline{0.290} & 
\underline{0.142} & 
\underline{0.272} & 
\underline{0.633} & 
\underline{0.754} &  0.331 & 
\underline{0.184} & 
\underline{0.201} & 
\underline{0.605} & 
0.731
   \\
ALM* (Ours)  & Y &  \textbf{0.271} & 
\textbf{0.132} & 
\textbf{0.302} & 
\textbf{0.699} & 
\textbf{0.808} &  \textbf{0.275} & 
\textbf{0.127} & 
\textbf{0.276} & 
\textbf{0.694} & 
\textbf{0.800}
 \\
\bottomrule
\end{tabular}
}
\vspace{-8mm}
\label{tab:image_inpainting_cmp_supp}
\end{table}

\vspace{-5mm}

\clearpage

\begin{figure}[h!]
	\centering
	\includegraphics[width=1.0\linewidth]{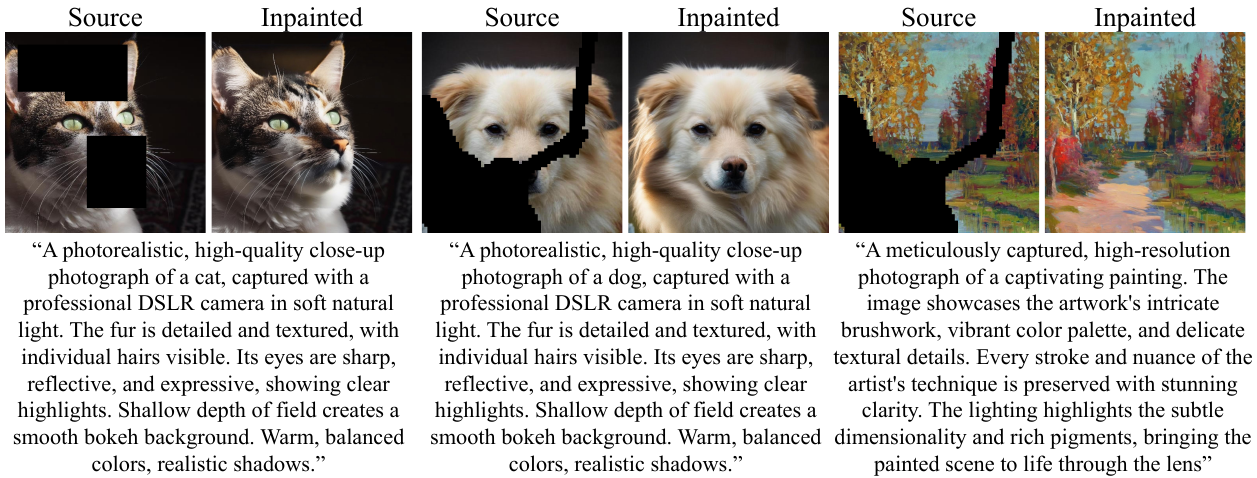}
	\vspace{-4mm}
	\caption{Qualitative results of ALM on complex scenes generated from extremely long and complex source prompts. We use Stable Diffusion XL~\cite{podell2023sdxl} for image generation.}
	\vspace{-8mm}
\label{fig:cplx_image}
\end{figure}

\paragraph{Results under complex text prompts.}
We demonstrate that our method maintains strong performance even when the source prompt is extremely long and packed with semantic detail, as shown in Figure~\ref{fig:cplx_image}.
For this experiment, the pretrained Stable Diffusion XL~\cite{podell2023sdxl} is used for image inpainting.

\vspace{-6mm}

\paragraph{Adaptation to flow matching frameworks.}

With flow matching frameworks~\cite{lipman2022flow, liu2022flow, flux2024}, we first modify the score estimation. 
Unlike diffusion models~\cite{sohl2015deep, ho2020denoising, song2020denoising} that predict the noise term $\epsilon_{\theta}(\by_t^i, t, \mathbf{c})$, flow matching models estimate the velocity field $v_{\theta}(\by_t^i, t, \mathbf{c})$. 
Accordingly, we reformulate the score estimation. 
In flow matching, the data point $\by_t^i$ along the probability path is defined as
\begin{equation}
	\by_t^i = (1-t)\by_0^i + t\epsilon.
	\label{eq:fm_path}
\end{equation}
Note that $\by_0^i$ denotes the clean image (\textit{i.e.}, real data), and $t=1$ corresponds to the pure noise endpoint $\epsilon$.
The corresponding velocity field is defined as
\begin{equation}
	v_{\theta}(\by_t^i,t,\mathbf{c}) = \epsilon - \by_0^i.
	\label{eq:fm_velocity}
\end{equation}
From Eq.~\eqref{eq:fm_path} and~\eqref{eq:fm_velocity}, we obtain $\by_0^i = \by_t^i - t v_{\theta}(\by_t^i,t,\mathbf{c})$.
Substituting this into Eq.~\eqref{eq:fm_path} yields
\begin{equation}
	\epsilon_{\theta}(\by_t^i,t,c) = \by_t^i + (1-t)v_{\theta}(\by_t^i,t,\mathbf{c}).
	\label{eq:epsilon_from_velocity}
\end{equation}
Using this relation, the score function is approximated as
\begin{equation}
	\nabla_{\by_t^i}\log p(\by_t^i|c) \approx
	-\frac{\epsilon_{\theta}(\by_t^i,t,c)}{\sigma_t} = - \frac{\by_t^i + (1-t)v_{\theta}(\by_t^i,t,\mathbf{c})}{\sigma_t},
	\label{eq:score_estimation_fm}
	\end{equation}
where $\sigma_t = t$ in flow matching frameworks.

Secondly, we adopt a different hyperparameter tuning strategy for flow-based models. 
Specifically, we set $w_1' = 0.5 w_1$ and use the following decaying schedule:
\begin{equation}
	w_i = \sigma_t^4 w_i^{\mathrm{init}}, \qquad \sigma_t = t.
\end{equation}
This modification is reasonable since flow matching models follow a fundamentally different sampling strategy compared to diffusion models. 
We leave the exploration of improved tuning strategies for flow-based frameworks as future work.

\begin{figure}[h!]
	\centering
	\includegraphics[width=1.0\linewidth]{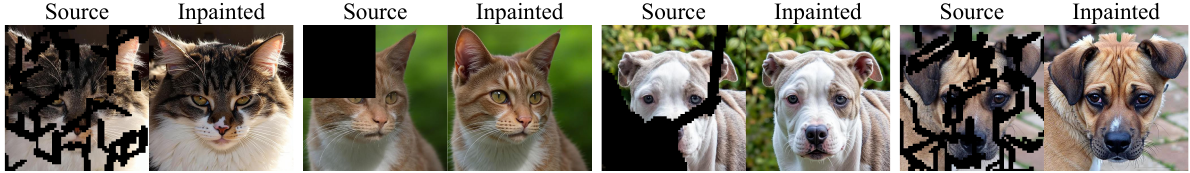}
	\vspace{-6mm}
	\caption{Qualitative results of ALM with the FLUX~\cite{flux2024} backbone for image inpainting on images sampled from the AFHQ dataset~\cite{choi2020stargan}. 
    These results demonstrate that ALM also performs well when combined with flow matching frameworks.
}
\label{fig:image_inpainting_flux_supple}
\end{figure}

\paragraph{Additional results with FLUX backbone.}
We first report the quantitative results obtained with the FLUX~\cite{flux2024} backbone in Table~\ref{tab:image_inpainting_flux}.
Note that the performance is comparable to the results obtained with other backbones (Stable Diffusion~\cite{rombach2022high} and Stable Diffusion XL~\cite{podell2023sdxl}), demonstrating that our method is model-agnostic.
We use 50 sampling steps for all experiments.
Additional qualitative results are visualized in Figure~\ref{fig:image_inpainting_flux_supple}.

\vspace{-4mm}
\begin{table}[h!]
\centering
\caption{Quantitative results of ALM across diverse backbones~\cite{rombach2022high, podell2023sdxl, flux2024}.}
\vspace{-3mm}
\setlength{\tabcolsep}{2.5pt} 
\scalebox{0.7}{
\begin{tabular}{l c c c c c c c c c c}
\toprule
\multirow{2}{*}{Method}  & \multicolumn{5}{c}{AFHQ~\cite{choi2020stargan}} & \multicolumn{5}{c}{CelebA-HQ~\cite{karras2017progressive}} \\
\cmidrule(lr){2-6} \cmidrule(lr){7-11}
& LPIPS $\downarrow$ & MSE $\downarrow$ & M-SSIM $\uparrow$ & MS-SSIM $\uparrow$ & FSIM $\uparrow$ & LPIPS $\downarrow$ & MSE $\downarrow$ & M-SSIM $\uparrow$ & MS-SSIM $\uparrow$ & FSIM $\uparrow$   \\
 \midrule
ALM (SD~\cite{rombach2022high})  &   {0.283} & {0.143} & {0.351} & {0.689} & {0.796}   &  {0.251} & {0.126}  & {0.417} & {0.732} & {0.813}  \\
ALM (SDXL~\cite{podell2023sdxl})  &  {0.254} & {0.112} &  {0.410} & {0.754} & {0.841} &  0.249 & {0.130} &  {0.442} & {0.772} & {0.843} \\
ALM (FLUX~\cite{flux2024})  &   0.330 & 0.124 & 0.277 & 0.649 & 0.761 &  0.302 & 0.152 & 0.298 & 0.670 & 0.763 \\
\bottomrule
\end{tabular}
}
\vspace{-12mm}
\label{tab:image_inpainting_flux}
\end{table}

\clearpage

\subsection{Wide Image Generation}

\begin{figure}[h!]
	\centering
	\includegraphics[width=0.9\linewidth]{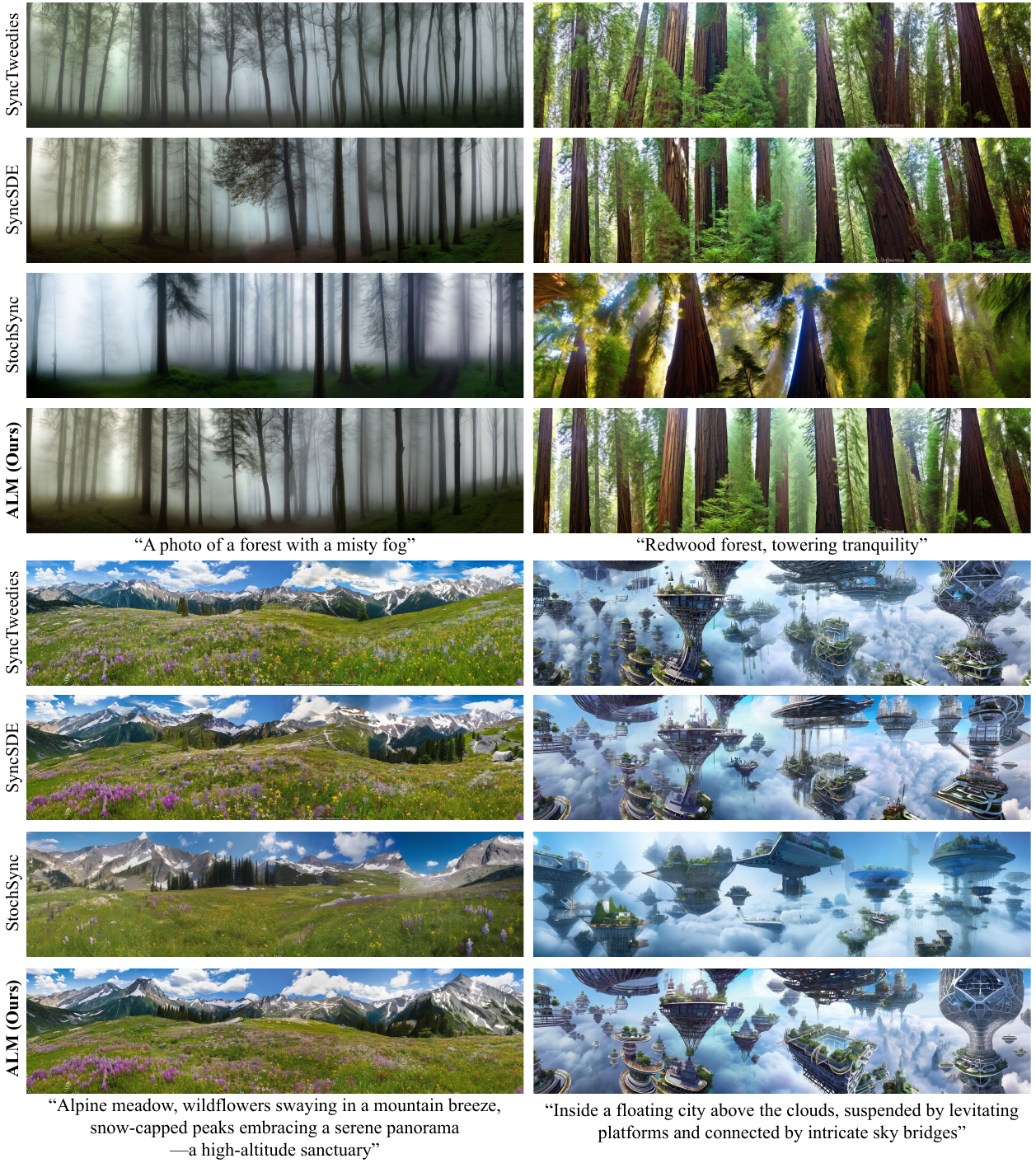}
	\vspace{-3mm}
	\caption{Additional qualitative comparison of our method with state-of-the-art methods~\cite{kim2024synctweedies,lee2025syncsde, yeo2025stochsync}.}
\label{fig:wide_image_supple}
\end{figure}

We use the pretrained Stable Diffusion~\cite{rombach2022high} v2.1-base model for the wide image generation task and employ a DDIM~\cite{song2020denoising} sampler with 50 timesteps.
For quantitative evaluation, we use a total of 8 prompts borrowed from prior works~\cite{bar2023multidiffusion, kim2024synctweedies, lee2025syncsde, yeo2025stochsync}.
Each prompt is used to generate 50 wide images with a resolution of $2048 \times 512$.
The generated wide images are then randomly cropped into $512 \times 512$ images for quantitative evaluation.
To compute FID~\cite{heusel2017gans} and KID~\cite{binkowski2018demystifying}, we generate 2,000 images per prompt using the same pretrained Stable Diffusion model and use them as a reference set.
Figure~\ref{fig:wide_image_supple} visualizes additional comparison of our method with baselines~\cite{kim2024synctweedies, lee2025syncsde, yeo2025stochsync} using diverse prompts. 
As shown, ALM effectively generates visually plausible images.

\vspace{-7mm}

\subsection{Human Motion Completion}

\begin{figure}[h!]
    \vspace{-8mm}
	\centering
	\includegraphics[width=0.85\linewidth]{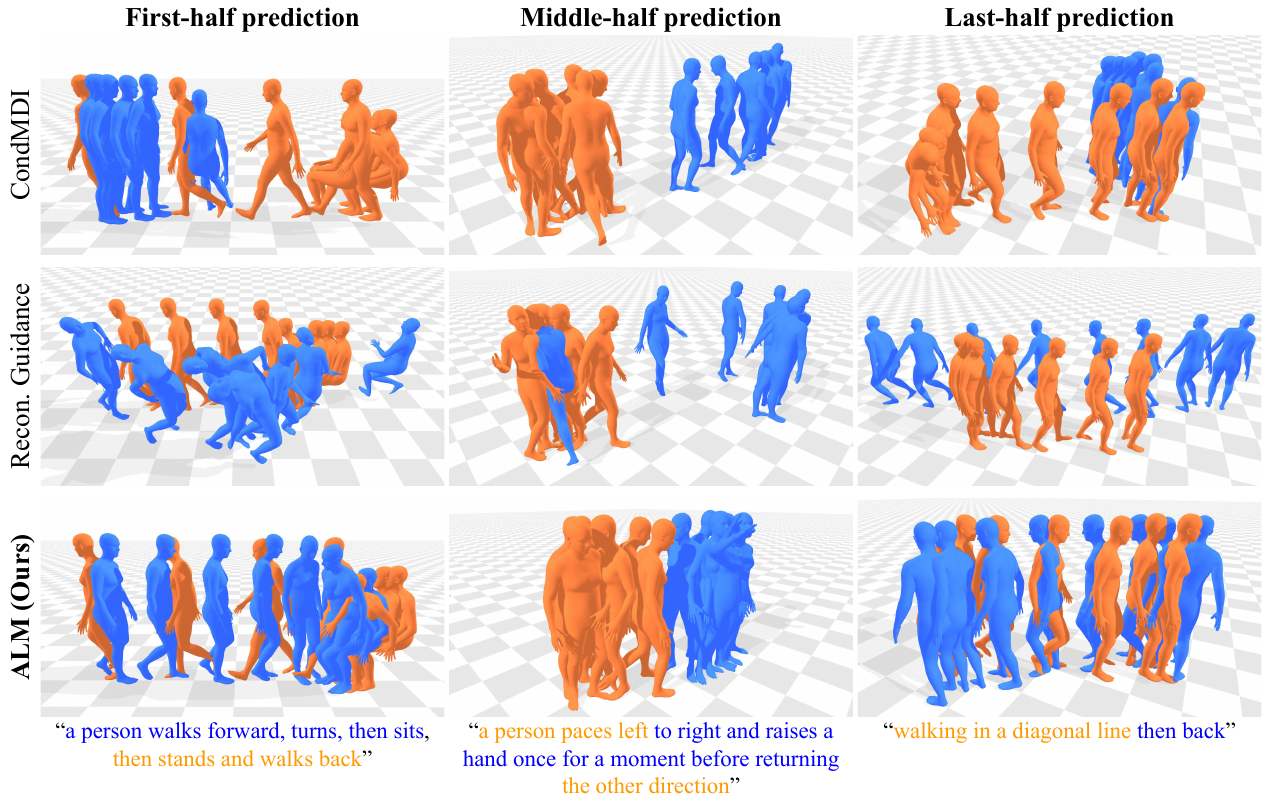}
	\vspace{-3mm}
	\caption{Additional qualitative comparison of our method with baselines~\cite{cohan2024flexible, ho2022video} on human motion completion. 
    While baselines show unrealistic or discontinuous motions, ALM generates plausible sequences that also align with the given text prompt.}
	\vspace{-9mm}
\label{fig:human_motion_completion_supple_1}
\end{figure}

\begin{figure}[h!]
	\centering
	\vspace{-5mm}
	\includegraphics[width=0.85\linewidth]{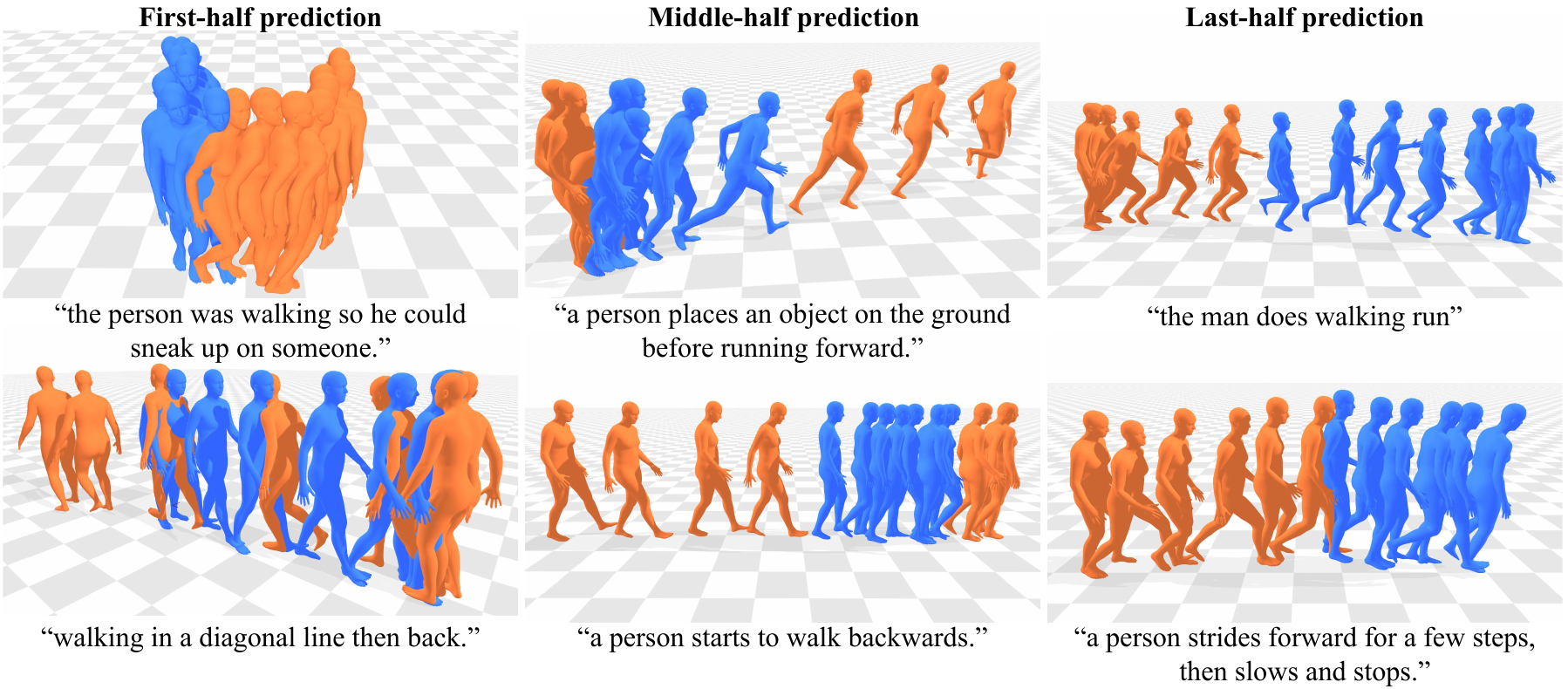}
	\vspace{-3mm}
	\caption{Additional qualitative results of human motion completion. 
    We show first-half (1st row), middle-half (2nd row) and last-half prediction scenario (3rd row).}
\label{fig:human_motion_completion_supple_2}
\end{figure}

Figures~\ref{fig:human_motion_completion_supple_1} and~\ref{fig:human_motion_completion_supple_2} illustrate additional qualitative results under three scenarios: 
``first-half prediction,'' ``middle-half prediction'', and ``last-half prediction,'' using the pretrained U-Net-based human motion diffusion model~\cite{karunratanakul2023guided}.
We visualize the given frames in \textcolor{orange}{orange}, and the synthesized frames in \textcolor{blue}{blue}. 
ALM effectively reconstructs the unobserved sequences of the human motion, while baselines~\cite{cohan2024flexible, ho2022video} struggle to generate realistic motions that align with the given text prompt.
Note that we follow the CondMDI setup, which employs a DDPM~\cite{ho2020denoising} sampler with 1,000 timesteps for motion sequence generation.
For this DDPM-based setting, we empirically modify the conditional distribution parameterization in Eq. (4) of the main paper, which improves performance:
\vspace{-1mm}
\begin{equation}
    p(\bx_t \mid \by_t, \mathbf{c}) := p(\bx_t \mid \by_t) \sim \mathcal{N}(\by_t, ({\sigma_t^2}/{w_1}) (\mathbf{1}-\bar{\bmask})^{-1}).
\end{equation}

\clearpage

\subsection{3D Mesh Texturing}

\begin{figure}[h!]
	\centering
	\includegraphics[width=1.0\linewidth]{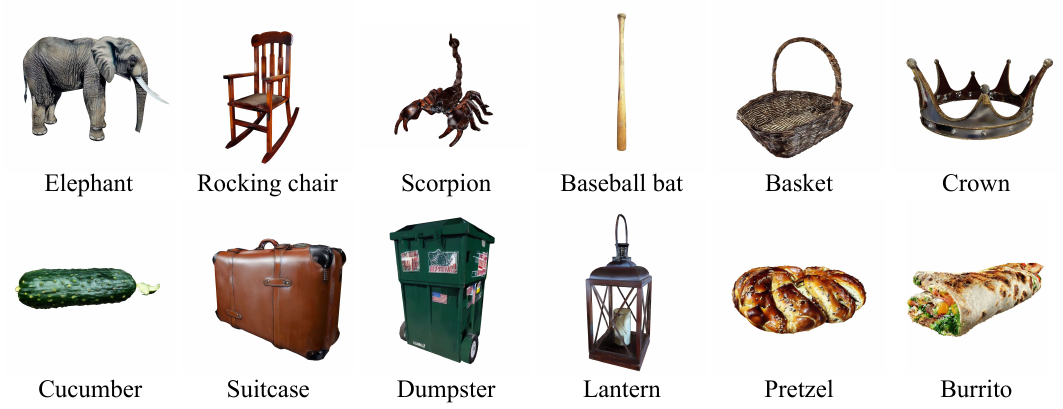}
	\vspace{-3mm}
	\caption{Additional qualitative results of 3D mesh texturing.}
\label{fig:mesh_texturing_supple}
\end{figure}

For 3D mesh texturing, we use the negative prompt ``oversmoothed, blurry, depth of field, out of focus, low quality, bloom, glowing effect.'' for all baselines. 
The overall experimental setup follows SyncTweedies~\cite{kim2024synctweedies} and SyncSDE~\cite{lee2025syncsde}.
During texturing, we sample a total of 10 viewpoints for the diffusion reverse process:
eight views are uniformly sampled over azimuth angles in $[0 \degree, 360 \degree)$ with an elevation of $0 \degree$, and two views are sampled at azimuths $0 \degree$ and $180 \degree$ with the elevation of $30 \degree$.
The same set of viewpoints is used for the quantitative evaluation.
The rendered images at each viewpoint are generated iteratively in an auto-regressive manner using the pretrained depth-conditioned ControlNet~\cite{zhang2023adding}. 
At the $i$-th viewpoint, mesh surface regions textured from the $1$st $\sim i-1$-th viewpoints are treated as pre-generated content, and the remaining untextured areas are interpreted as unobserved regions.
To obtain the rendered view of the textured region at $i$-th viewpoint using $i-1$ rendered images, we synthesize an auxiliary texture map from the rendered views of the $1$st $\sim i-1$-th viewpoints.
After sampling rendered images from $1$st $\sim 10$-th viewpoints, these images are used to generate the final texture map.
For texture baking, the texture map is optimized by minimizing the rendering loss over the 10 rendered images. 
During the generation of the final texture map, Voronoi diagram-based completion~\cite{aurenhammer1991voronoi} is applied. 
The RGB texture map has a resolution of $1024 \times 1024$, while the latent texture map has a resolution of $1536 \times 1536$.
The rendered images synthesized by the diffusion model have a resolution of $768 \times 768$.
We also apply the modified attention operation used in prior works~\cite{kim2024synctweedies, liu2024text} during diffusion sampling to ensure better results.
We use a total of 30 DDIM~\cite{song2020denoising} sampling steps and set $\dby_t = \mathbf{0}$ during the final 20\% of the reverse process to ensure a fair comparison with prior works~\cite{kim2024synctweedies, lee2025syncsde}.
PyTorch3D~\cite{ravi2020pytorch3d} and Nvdiffrast rasterizer~\cite{Laine2020diffrast} are used for mesh rendering.
Figure~\ref{fig:mesh_texturing_supple} shows additional qualitative results of ALM on the 3D mesh texturing task.

\vspace{-4mm}

\clearpage

\subsection{Long Video Generation}

\begin{figure}[h!]
\vspace{-8mm}
	\centering
	\includegraphics[width=1.0\linewidth]{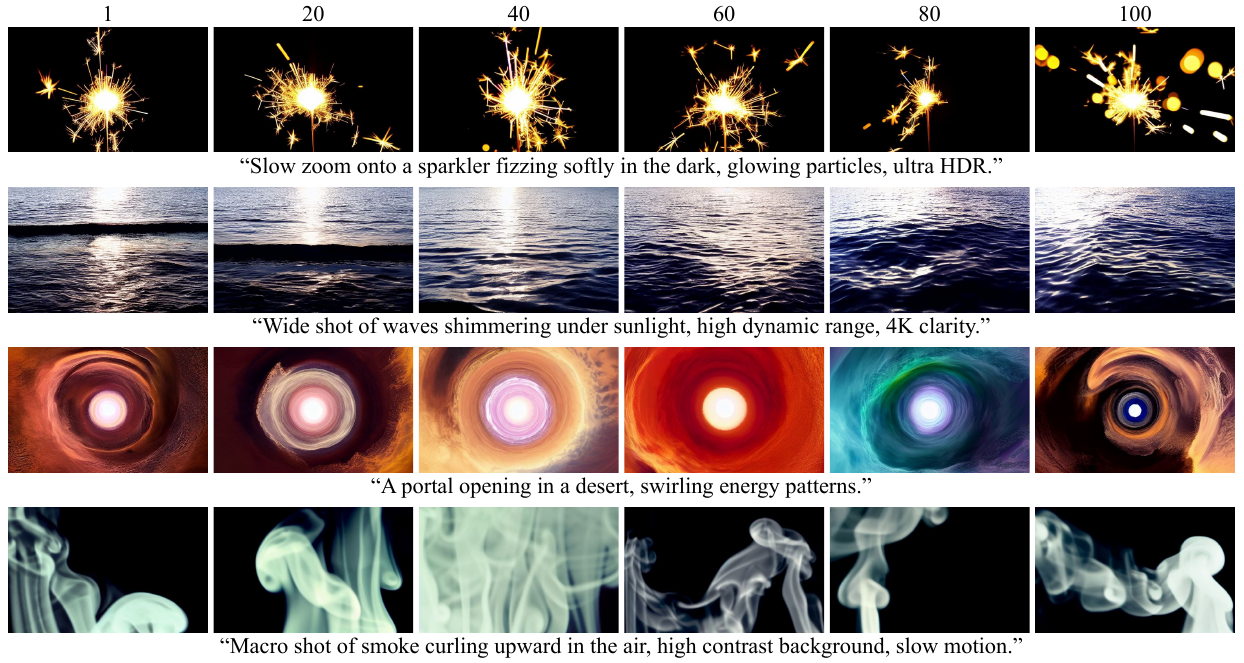}
	\vspace{-6mm}
	\caption{Qualitative results of long video generation. 
    We use the pretrained LaVie~\cite{wang2025lavie}, which generates 16 frame videos by default, and extend the synthesized videos to 104 frames using ALM.}
	\label{fig:long_video}
\end{figure}

We extend our framework into the video domain by generating temporally long sequences. 
Analogous to the wide image generation task, we produce video frames by autoregressively sampling multiple overlapping video patches along the temporal axis. 
For implementation, we use the pretrained LaVie~\cite{wang2025lavie}, a diffusion-based text-to-video model that generates $512 \times 320$ resolution videos in the first stage, producing 16-frame videos from a single text prompt. 
By setting a temporal stride of 8 and synthesizing a total of 12 patches, we generate a 104-frame video with an FPS of 8.
Here, we use a DDIM~\cite{song2020denoising} sampler with 50 steps.
Consecutive patches are overlaid along the temporal axis and decoded using LaVie's pretrained VAE~\cite{kingma2013auto} decoder.
We visualize the generated long video sequences in Figure~\ref{fig:long_video}.
As shown, ALM produces visually coherent and semantically consistent sequences, maintaining spatio-temporal continuity.

For quantitative evaluation, we compare our method with three baselines:
FreeNoise~\cite{qiu2023freenoise}, SEINE~\cite{chen2023seine}, and SyncSDE~\cite{lee2025syncsde}.
A total of 261 video sequences with diverse text prompts are used for evaluation, each containing 104 frames.
The text prompts are either borrowed from prior work~\cite{wang2025lavie} or generated using ChatGPT.
Evaluation metrics include FVD~\cite{unterthiner2019fvd}, KVD~\cite{unterthiner2019fvd}, and CLIP text-video similarity~\cite{radford2021learning} (CLIP-Sim).
For FVD and KVD computation, the reference videos are generated using the pretrained LaVie. 
The generated long video from each method is split into 12 sub-videos of 16 frames using a temporal stride of 8, and these sub-videos are used for evaluation.
For CLIP text-video similarity, the image--prompt similarity is computed for each of the 104 frames and then averaged.
Quantitative results are presented in Table~\ref{tab:long_video_quant}.
As shown, our method outperforms the baselines, emphasizing the effectiveness of our method.
Note that we scale the value of FVD and KVD by $10^{-3}$.

\begin{table}[h!]
\centering
\caption{Quantitative comparison of ALM with baselines~\cite{qiu2023freenoise, chen2023seine, lee2025syncsde} on long video generation task.
FVD~\cite{unterthiner2019fvd} and KVD~\cite{unterthiner2019fvd} values are scaled by $10^{-3}$.}
\vspace{-3mm}
\setlength{\tabcolsep}{8pt}
\scalebox{0.95}{
\begin{tabular}{lccc}
\toprule
Method & FVD $\downarrow$ & KVD $\downarrow$ & CLIP-Sim $\uparrow$ \\
\midrule
FreeNoise~\cite{qiu2023freenoise} & 2.404 & \underline{3.119} & \underline{0.312} \\
SEINE~\cite{chen2023seine} & 3.501 & 4.219 & 0.305 \\
SyncSDE~\cite{lee2025syncsde} & \underline{2.290} & 3.135 & \underline{0.312} \\
ALM (Ours) & \textbf{2.215} & \textbf{2.859} & \textbf{0.313} \\
\bottomrule
\end{tabular}
}
\vspace{-9mm}
\label{tab:long_video_quant}
\end{table}

\section{Discussion on Score Estimation}
\label{sec:supp_score_estimation}

In this section, we empirically justify the score estimate $\nabla_{\by_t^i} \log p(\bx_t, \bmask, \by_t^i \mid \mathbf{c}) \simeq \nabla_{\by_t^i} \log p(\be_t^i \mid \mathbf{c})$ introduced in Sec. 3.3 of the main paper.
Specifically, we investigate whether the pretrained diffusion model inherently captures the joint structural coherence dictated by the mask $\bmask$.
A robust estimator of the true joint score should be fundamentally invariant to minor morphological variations in $\bmask$. 
The semantic and structural compatibility between the pre-generated content $\bx_t$ and the unobserved region $\by_t^i$, represented by $\be_t^i$, relies on the global context rather than the exact pixel-wise precision of their boundary.
Assuming the score function $\mathbf{s}(\be_t^i \mid \mathbf{c}) = \nabla_{\by_t^i} \log p(\be_t^i \mid \mathbf{c})$ evaluates this joint coherence, it should remain highly consistent when the mask geometry is slightly modified. 
Alternatively, a score that merely penalizes shallow boundary artifacts would exhibit severe fluctuations under such structural perturbations.

To verify this robustness, we introduce two distinct degradations to $\bmask$, generating a perturbed mask $\bmask'$ and a correspondingly degraded composition $\be_t^{i'} = \bx_t \odot (\mathbf{1} - \bmask') + \by_t^i \odot \bmask'$. 
These degradations induce local boundary perturbations without altering the global context of the scene. 
We employ two distinct operations:
(a) Dilation: applying morphological dilation to artificially alter the strict structural boundary between the observed and unobserved regions, and 
(b) Boundary noise addition: injecting Gaussian noise along the mask edges.

\begin{figure}[h!]
	\centering
	\includegraphics[width=1.0\linewidth]{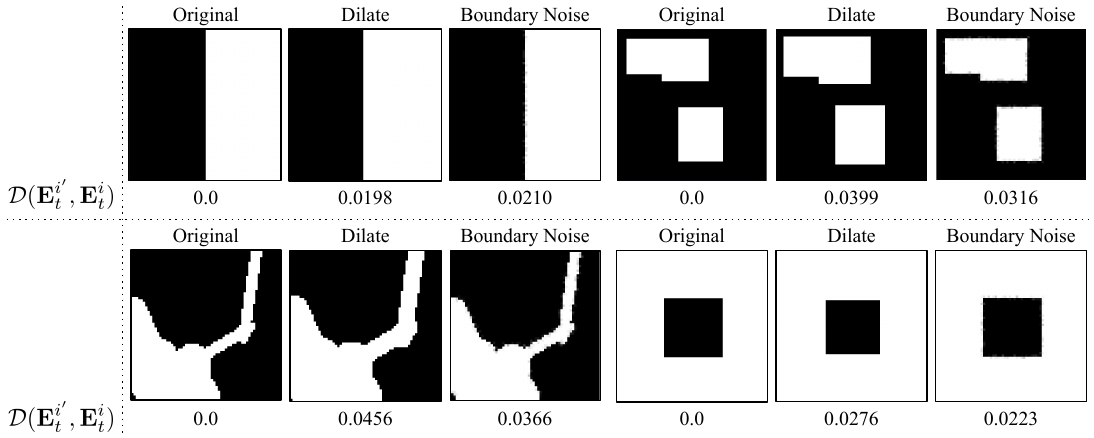}
    \vspace{-4mm}
	\caption{Empirical justification of the score estimation of $\nabla_{\by_t^i} \log p(\bx_t, \bmask, \by_t^i \mid \mathbf{c})$. 
    We visualize the degraded masks, and report the measured deviation.
    Degraded masks are best visualized when magnified.
    As shown, the deviation remains small, verifying our score estimation strategy.
    }
    \vspace{-4mm}
\label{fig:score_justification}
\end{figure}

To quantify the sensitivity of the score estimation, we measure the relative deviation between the original and perturbed scores:
\begin{equation}
    \mathcal{D}(\be_t^{i'}, \be_t^i) = \mathbb{E}_{t, \bx_t, \by_t^i} \left[ \frac{\| \mathbf{s}(\be_t^{i'} \mid \mathbf{c}) - \mathbf{s}(\be_t^i \mid \mathbf{c}) \|}{\| \mathbf{s}(\be_t^i \mid \mathbf{c}) \|} \right].
\end{equation}
As shown in Figure~\ref{fig:score_justification}, the deviation $\mathcal{D}(\be_t^{i'}, \be_t^i)$ remains remarkably small despite the applied degradations and varying mask geometries. 
This minimal deviation confirms that the score estimation $\mathbf{s}(\be_t^i \mid \mathbf{c})$ is not overly sensitive to local mask artifacts. 
Consequently, the score of $p(\be_t^i \mid \mathbf{c})$ serves as a robust surrogate for estimating the score of the true joint distribution $p(\bx_t, \bmask, \by_t^i \mid \mathbf{c})$.

This robustness is also consistent with prior diffusion applications that use pretrained models on locally blended variables.
For example, blended latents are used in image inpainting~\cite{avrahami2023blended} and motion imputation~\cite{tevet2022human}, where pretrained diffusion models still produce strong results.
Similarly, we observe stable performance across diverse backbones and tasks, suggesting that the pretrained score network can provide a useful estimate for the blended variable $\be_t^i$ in practice.

\section{Discussion on Acceleration Strategy}
\label{sec:supp_acceleration}

\begin{figure}[h!]
	\centering
	\includegraphics[width=0.95\linewidth]{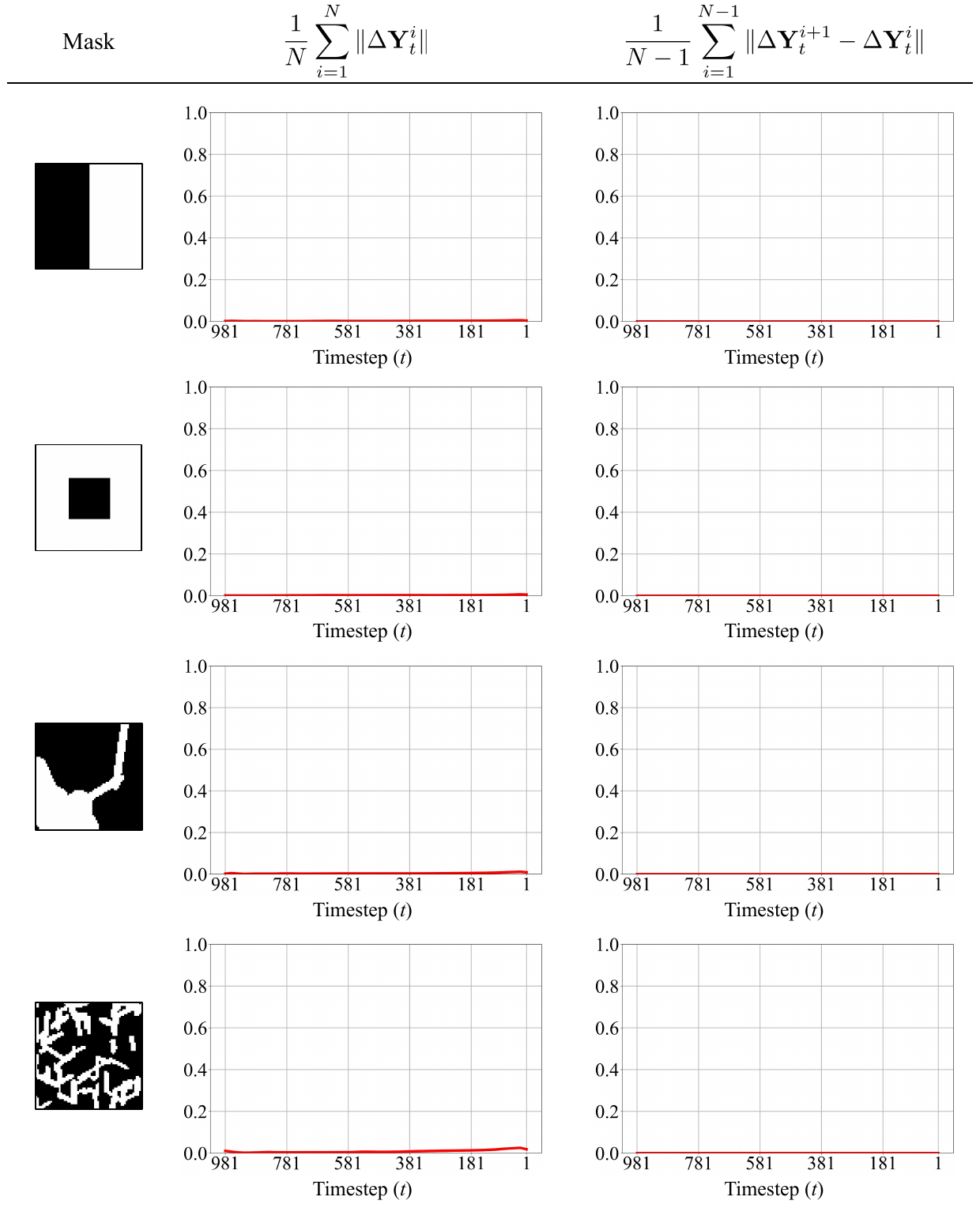}
    \vspace{-3mm}
	\caption{Analysis on the acceleration strategy.
    We show that the norm of $\dby_t^i$ remains sufficiently small (Column 2), and the proposed one-step approximation is also valid (Column 3).
    These results justify the one-step approximation and support the validity of our acceleration strategy and its underlying assumptions. 
    }
\label{fig:acc_ablation}
\end{figure}

In this section, we validate the one-step approximation strategy described in Section 3.4 of the main paper. 
Our key claim is the following: when Claim 1 holds, \textit{i.e.}, $\|\Delta \mathbf{Y}_t^{i}\| \ll 1$, the difference between consecutive updates also remains small, that is, $\|\Delta \mathbf{Y}_t^{i+1} - \Delta \mathbf{Y}_t^{i}\| \ll 1$.
To verify this, we sample the sequence $\{\Delta \mathbf{Y}_t^i\}_{i=1}^{N}$ {without using the acceleration strategy} and check whether our claim holds. 
We set $N=500$.
In practice, the coefficients $\lambda_1$ and $\lambda_2$ are chosen to ensure that $\| \dby_t^i\|$ remains sufficiently small, thereby satisfying Claim 1. 
In our experiments, we use $\lambda_1 = 2 \times 10^{-3}$ and $\lambda_2 = 10^{-5}$, which correspond to the configuration $w_1=1$ and $w_2=0.005$ used in the image inpainting experiments of the main paper. 
Therefore, Claim 1 is not merely an assumption but is enforced by the choice of coefficients in our practical setup.
Figure~\ref{fig:acc_ablation} shows the experimental results under various mask geometries. 
In Column 2, we visualize the average norm of $\dby_t^i$ across iterations for each timestep $t$, defined as:
\begin{equation}
    \frac{1}{N}\sum_{i=1}^{N}\|\Delta \mathbf{Y}_t^{i}\|,
\end{equation}
which empirically confirms that Claim 1 is valid.
Next, Column 3 plots the average difference between $\dby_t^{i+1}$ and $\dby_t^i$ for each timestep $t$:
\begin{equation}
    \frac{1}{N-1}\sum_{i=1}^{N-1}\|\Delta \mathbf{Y}_t^{i+1}-\Delta \mathbf{Y}_t^{i}\|.
    \label{eq:update_difference}
\end{equation}
As shown, the value of Eq.~\eqref{eq:update_difference} is small enough, thereby validating the proposed one-step approximation regardless of the mask shape.
Since the acceleration strategy is applied on top of this configuration that already satisfies Claim 1, the above results justify the use of the one-step approximation in our method. 
Notably, this approximation dramatically reduces runtime ($185\times$) as reported in Table~\ref{tab:computation}, while maintaining performance as demonstrated in Table 3 of the main paper, highlighting its effectiveness.

\vspace{-6mm}

\section{Analysis on Computational Cost}
\label{sec:supp_computation}

We quantitatively evaluate the computational cost of ALM and the baselines~\cite{ju2024brushnet, zhuang2024task, rombach2022high, lee2025syncsde, manukyan2023hd} on the image inpainting task.
Table~\ref{tab:computation} reports the required GPU memory and runtime for generating a single image. 
Overall, ALM is fully training-free, and its computational cost remains on par with existing training-free baselines.
These findings highlight that ALM achieves the superior trade-off between performance and computational efficiency among the various methods.

\vspace{-5mm}
\begin{table}[h!]
\centering
\caption{{Quantitative computational cost analysis across diverse baselines.}}
\vspace{-3mm}
\setlength{\tabcolsep}{6pt}
\scalebox{0.85}{
\begin{tabular}{lcccc}
\toprule
Method & Training-Free & \begin{tabular}[c]{@{}c@{}}GPU Memory (GB) $\downarrow$ \end{tabular} & \begin{tabular}[c]{@{}c@{}}Runtime (s) $\downarrow$ \end{tabular} & LPIPS $\downarrow$ \\
\midrule
BrushNet~\cite{ju2024brushnet}      & N & 4.73 & {3.189}  & 0.316 \\
PowerPaint~\cite{zhuang2024task}    & N    & 5.54 & 4.089  & 0.310 \\
SDI~\cite{rombach2022high}     & N    & 3.28 & 2.223  & 0.292 \\
SyncSDE~\cite{lee2025syncsde}   & Y  & 4.98 & 6.734  & 0.304 \\
HD-Painter~\cite{manukyan2023hd}  & Y    & 29.07 & 38.840 & 0.301 \\
ALM (Ours)    & Y  & 4.98 & 9.900 & 0.283 \\
ALM w/o Acceleration   & Y & 4.98 & 1854.042  & 0.298 \\
\bottomrule
\end{tabular}
}
\label{tab:computation}
\vspace{-8mm}
\end{table}

\section{Analysis on Hyperparameter Sensitivity}
\label{sec:supp_hyp}

\begin{figure}[ht!]
\vspace{-5mm}
	\centering
	\includegraphics[width=1.0\linewidth]{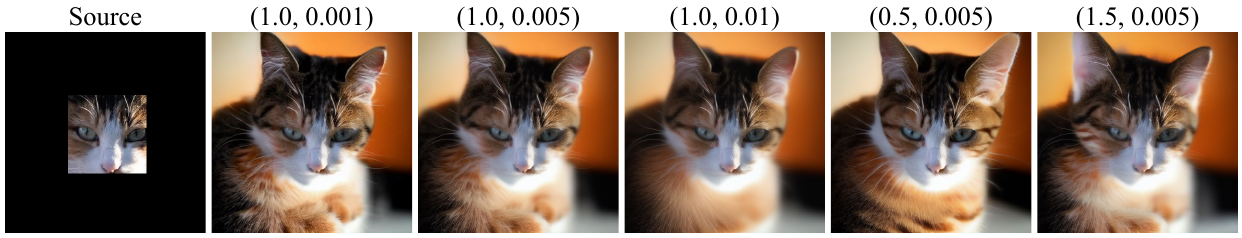}
	\vspace{-6mm}
	\caption{Qualitative results of ALM under five hyperparameter settings ($w_1, w_2$). Our method is robust to hyperparameter configurations.}
\label{fig:hyp_sensitivity}
\end{figure}

As mentioned in Sec. 3.4, we use two hyperparameters: $w_1$ and $w_2$.
We now demonstrate that ALM is robust under variations of these hyperparameters through additional experiments conducted on the image inpainting task.
We sweep $w_1$ over $[0.5, 1.0, 1.5]$ and $w_2$ over $[0.001, 0.005, 0.01]$, then provide the corresponding quantitative results in Table~\ref{tab:hyp_sensitivity} as well as qualitative comparisons in Figure~\ref{fig:hyp_sensitivity}.
As shown, our method consistently maintains strong performance across all tested configurations, confirming the robustness of the proposed method.

\vspace{-5mm}
\begin{table}[h!]
\centering
\caption{Quantitative analysis of hyperparameter sensitivity on image inpainting.}
\vspace{-3mm}
\setlength{\tabcolsep}{3pt} 
\scalebox{0.63}{
\begin{tabular}{l c c c c c c c c c c}
\toprule
\multirow{2}{*}{Method} & \multicolumn{5}{c}{AFHQ~\cite{choi2020stargan}} & \multicolumn{5}{c}{CelebA-HQ~\cite{karras2017progressive}} \\
\cmidrule(lr){2-6} \cmidrule(lr){7-11}
 & LPIPS $\downarrow$ & MSE $\downarrow$ & M-SSIM $\uparrow$ & MS-SSIM $\uparrow$ & FSIM $\uparrow$ & LPIPS $\downarrow$ & MSE $\downarrow$ & M-SSIM $\uparrow$ & MS-SSIM $\uparrow$ & FSIM $\uparrow$   \\
\midrule
Baseline (Best)     &  0.304 & 0.172 & {0.302} & {0.641} & {0.778} & {0.268} & {0.130} & {0.368} & 0.659 & 0.763 \\
\hdashline
ALM ($w_1{=}1.0$, $w_2{=}0.001$)  &  \underline{0.283} &  0.147 &  0.333 &  0.681 &  0.794 &  0.253 & 0.130 & 0.392 & 0.723 & 0.809 \\
ALM ($w_1{=}1.0$, $w_2{=}0.005$)  &  \underline{0.283} & 0.143 & 0.351 & 0.689 & \underline{0.796} &  \underline{0.251} &  0.126 &  0.417 &  0.732 &  \underline{0.813} \\
ALM ($w_1{=}1.0$, $w_2{=}0.01$)   &   0.288 & \underline{0.139} & \textbf{0.367} & \textbf{0.694} & 0.795 &  0.255 &  \underline{0.124} &  \textbf{0.436} &  \textbf{0.738} &  \underline{0.813} \\
ALM ($w_1{=}0.5$, $w_2{=}0.005$)  & 0.292 & 0.158 & 0.342 & 0.675 & 0.791 &  0.264 &  0.138 &  0.412 &  0.719 &  0.804 \\
ALM ($w_1{=}1.5$, $w_2{=}0.005$)  &  \textbf{0.281} & \textbf{0.137} & \underline{0.356} & \underline{0.692} & \textbf{0.798} &  \textbf{0.250} & \textbf{0.122} & \underline{0.418} & \underline{0.734} & \textbf{0.815} \\
\bottomrule
\end{tabular}
}
\vspace{-8mm}
\label{tab:hyp_sensitivity}
\end{table}

\end{document}